\documentclass[journal]{IEEEtran}

\usepackage{physics}
\usepackage{amsmath,amssymb}
\usepackage{mathtools,bbm,bm}
\usepackage{graphicx}
\usepackage{aligned-overset}
\usepackage{xcolor}
\usepackage{algorithm, algpseudocode}
\usepackage{hyperref,cleveref}
\usepackage{lipsum}
\usepackage{orcidlink}

\usepackage{tcolorbox}
\newtcolorbox{mymathbox}[1][]{colframe=black!15!, coltitle=black, colback=white, sharp corners, #1}

\usepackage{subcaption}

\newcommand{\mycomment}[1]{}

\newcommand{\ParR}[1]{\left(#1\right)}
\newcommand{\ParE}[1]{\left[#1\right]}

\newcommand{\wt}[1]{\widetilde{#1}}

\newcommand{\muilam}[2]{(#1/#1_I,#2)}

\newcommand{\rot}[1]{{\color{red}#1}}

\newcommand{\SUMM}[0]{\sum_{m=1}^{\mu}}

\newcommand{\CMULAM}[0]{c_{\mu/\mu,\lambda}}
\newcommand{\CTHETA}[0]{c_{\vartheta}}
\newcommand{\CTHETAlong}[0]{\ONETWOPI \frac{1}{\vartheta} \EXP{ -\frac{1}{2} \ParR{\Phi^{-1}(\vartheta)}^2 }}

\newcommand{\ONETWOPI}[1][]{\frac{1}{\sqrt{2\pi}#1}}

\newcommand{\EXP}[1]{\operatorname{exp}\left[ #1 \right]}

\newcommand{\BigO}[1]{O\ParR{#1}}

\newcommand{\ceil}[1]{\lceil #1 \rceil}
\newcommand{\floor}[1]{\lfloor #1 \rfloor}

\newcommand{\EV}[1]{\operatorname{E}\left[#1\right]}
\newcommand{\VAR}[1]{\operatorname{Var}\left[#1\right]}

\newcommand{\vt}[0]{\vartheta}

\newcommand{\signss}[0]{\sigma^*_{\mathrm{ss}}}

\newcommand{\signzero}[0]{\sigma^*_{\varphi_0}}

\newcommand{\signopt}[0]{\hat{\sigma}^*}

\newcommand{\rec}[1]{\langle#1\rangle}

\newcommand{\cs}[0]{c_\sigma}
\newcommand{\gvec}[2]{\vb{#1}^{(#2)}}

\newcommand{\fmed}[1]{f_\mathrm{med}^{(#1)}}
\newcommand{\ppt}[0]{\norm{\vb{p}_\theta}^2}
\newcommand{\ppc}[0]{\norm{\vb{p}_c}^2}
\newcommand{\ppm}[0]{\norm{\vb{p}_m}^2}
\newcommand{\gmc}[2]{#1^{(#2)}_{\theta}}
\newcommand{\gmcm}[2]{#1^{(#2)}_{m}}
\newcommand{\gmcc}[2]{#1^{(#2)}_{c}}
\newcommand{\dm}[0]{\Delta\vb{m}}
\newcommand{\dS}[0]{\Delta\bm{\Sigma}}
\newcommand{\ratiog}[1]{\frac{#1^{(g+1)}}{#1^{(g)}}}

\newcommand{\Per}[0]{\mathcal{P}}
\newcommand{\Thr}[0]{\mathcal{T}}

\newcommand{\mumax}[0]{\mu_\mathrm{max}}
\newcommand{\rsigsqrt}[0]{\sqrt{\mu^{(g+1)}/\mu^{(g)}}}

\hypersetup{colorlinks=true, citecolor=black, linkcolor=black, urlcolor=black}
\crefformat{equation}{(#2#1#3)}
\crefformat{section}{#2#1#3}
\crefformat{figure}{#2#1#3}
\crefformat{algorithm}{#2#1#3}
\crefformat{table}{#2#1#3}
\crefformat{appendix}{#2#1#3}
\crefformat{line}{#2#1#3}

\newcommand{\mywidth}{4.3cm}
\begin{document}
	
	\title{On the Interaction of Adaptive Population Control with Cumulative Step-Size Adaptation}
	%
	%\author{
			%\IEEEauthorblockN{Amir Omeradzic\textsuperscript{\orcidlink{0000-0003-1979-8916}} and Hans-Georg Beyer\textsuperscript{\orcidlink{0000-0002-7455-8686}}}%\\
			%\IEEEauthorblockA{Vorarlberg University of Applied Sciences, Research Center Business Informatics
				%\\\{amir.omeradzic, %hans-georg.beyer\}@fhv.at}
	
	\author{\IEEEauthorblockN{Amir Omeradzic and Hans-Georg Beyer}
	\thanks{The authors are with the Research Center Business Informatics, Vorarlberg University of Applied Sciences, 6850 Dornbirn, Austria.
	
	This research was funded by the Austrian Science Fund (FWF) under grant P33702-N.
	
	This work is licensed under a \href{https://creativecommons.org/licenses/by/4.0/}{Creative Commons Attribution 4.0 License}.
	}
	}

	\maketitle
	\mycomment{
	\IEEEpubid{\begin{minipage}{\textwidth}\ \\[12pt] \centering
		This work is licensed under a \href{https://creativecommons.org/licenses/by/4.0/}{Creative Commons Attribution 4.0 License}
	\end{minipage}} 
	\IEEEpubidadjcol %space in 1st column
	}
	
	\begin{abstract}
			Three state-of-the-art adaptive population control strategies (PCS) are theoretically and empirically investigated for a multi-recombinative, cumulative step-size adaptation Evolution Strategy $\muilam{\mu}{\lambda}$-CSA-ES.
			First, scaling properties for the generation number and mutation strength rescaling are derived on the sphere in the limit of large population sizes.
			Then, the adaptation properties of three standard CSA-variants are studied as a function of the population size and dimensionality, and compared to the predicted scaling results.
			Thereafter, three PCS are implemented along the CSA-ES and studied on a test bed of sphere, random, and Rastrigin functions.
			The CSA-adaptation properties significantly influence the performance of the PCS, which is shown in more detail.
			Given the test bed, well-performing parameter sets (in terms of scaling, efficiency, and success rate) for both the CSA- and PCS-subroutines are identified.
			%Furthermore, strengths and weaknesses of each PCS-method are discussed.
	\end{abstract}
	
	\section{Introduction}
\label{sec:intro}
%\lipsum[1-3]

Evolution Strategies (ES) have shown to be well-suited for the optimization of test functions subjected to strong noise or high multimodality with adequate global structure.
The crucial strategy parameter in both cases is the population size.
On noisy functions, such as the noisy sphere, large populations enable to improve the signal-to-noise ratio of the ES and reduce the expected residual distance to the optimizer \cite{AB00d}.
For well-structured highly multimodal problems, e.g., the Rastrigin function, they help to facilitate a global search, enabling to locate the global optimizer among exponentially many (in the dimensionality) local optima \cite{SB23,OB23}.
While models for the population sizing for an multi-recombinative ES on the Rastrigin function are derived in \cite{SB23,OB23}, the results of \cite{OB24} demonstrate the importance of tuning the population size for the highest efficiency on Rastrigin.

Adaptive population control strategies (abbreviated as PCS) are able to adapt to both strong noise \cite{BS06a,HB16,NA18} and high multimodality \cite{NA18,NH17} by increasing the population size when insufficient algorithm performance is detected.
While benchmarks on the well-known Black-Box Optimization Benchmarking (BBOB) test bed \cite{hansen2021coco} are available in many cases,
an in-depth analysis of adaptive population control is still pending.
Furthermore, PCS are often tuned for the underlying test bed, such that direct comparison between different methods turns out to be difficult in some cases.
In principle, adaptive population control can be implemented into any ES.
State-of-the-art ES use covariance matrix adaptation (CMA) \cite{hansen2023cma} for the search space distribution together with cumulative step-size adaptation (CSA) \cite{HO01,Arn02} for the global mutation strength.
As it turns out, there is significant dependence of the PCS performance on the underlying CMA- and CSA-parametrization.
This also holds among standard implementations of the CSA-ES, which will be shown.
Population control cannot be studied isolated from the underlying mutation strength adaptation.
A change of the population size usually results in changes of the mutation strength, which in turn influences the performance measure of the PCS.
Hence, three state-of-the-art PCS routines \cite{HB16,NA18,NH17} will be selected that are suitable for both noisy and multimodal problems.
The PCS will incorporate performance measures both in search space and in fitness space, respectively.
However, they will be investigated for isotropic mutations by using only the CSA-ES (without CMA) on a simple test bed.
The idea is to reduce the problem complexity and provide theoretical reasoning for the parameter choices.
Hence, adaptation time-scales and mutation strength rescaling are analyzed for CSA and PCS both theoretically and experimentally.
The obtained parametrization is then applied to all PCS, enabling a direct comparison of the methods for the first time.
Furthermore, the influence of different CSA-variants on adaptive population control is illustrated.

%An analysis of CSA has been done in \cite{Arn02} for small populations and large dimensionality on the sphere function.
%However, for adaptive population control, a different approach is needed since large populations at (comparably) small dimensionality occur.
%Therefore, the present paper will provide an extension of the analysis from \cite{Arn02} by assuming large populations.

In the next Sec.~\cref{sec:csa}, the adaptive population control routine is introduced, which is later investigated in Sec.~\cref{sec:pcs}.
In Sec.~\cref{sec:csa_analysis}, preliminary theoretical results are derived which enable a better understanding the ES in the limit of large population sizes.
In Sec.~\cref{sec:schedule}, a simple population change schedule is introduced to test the stability of the CSA.
Then, the adaptive PCS are introduced and discussed in Sec.~\cref{sec:pcs}.
Furthermore, their performances are compared and the influence of the CSA is discussed.
Finally, conclusions are drawn in Sec.~\cref{sec:conc}.

%The PCS investigated later in this paper use covariance matrix adaptation

\mycomment{

There are two main goals of the present paper.

Both adaptive population control and the CSA-ES aggregate information over certain time scales, i.e., over a number of generation.
The relevant time scales usually scale with the search space dimensionality

The covariance matrix adaptation CMA-ES with cumulative step-size adaptation (CSA) for the (global) mutation strength adaptation constitutes a state-of-the-art algorithm for global optimization.
Furthermore, there are numerous PCS implementations based on modifications of the default CMA-ES (e.g. \cite{NH17,NA18}) and benchmarks on the BBOB test bed \rot{??}.
However, it is clear that a theoretical investiagation of adaptive popualtion

, theoretical analysis in its infancy.
A recent study of self-adaptation on the Rastrigin function \cite{OB24} has investigated the 

\begin{itemize}
	\item motivate population control
	\item motivate usage of sphere as simplified model
	\item PCS
	\item Schrittweitenadaptation in der Evolutionsstrategie mit einem
	entstochastisierten Ansatz. Dissertation, Ostermeier
\end{itemize}
}
	\section{CSA-ES with Adaptive Population Control}
\label{sec:csa}

The multi-recombinative $\muilam{\mu}{\lambda}$-CSA-ES with adaptive population control is given in Alg.~\cref{alg:pcs_algos}.
It operates with isotropic mutations of strength $\sigma$.
At this point, the goal is to introduce the general algorithm, while the details of the respective population control subroutines are presented later in Alg.~\cref{alg:pcs_subr}.
The parent and offspring population is denoted by $\mu$ and $\lambda$, respectively, with the truncation ratio $\vartheta \coloneqq \mu/\lambda$ ($\vt=1/2$ will be used).
The CSA-specific parameters (Lines 2, 7-17) are discussed in Sec.~\cref{sec:csa_analysis}.
Generic (common) parameters of the PCS are initialized in Line 3 and specific parameters in Line 4.
The PCS-methods to be used are based on \cite{NH17} (APOP), \cite{HB16} (pcCSA), and \cite{NA18} (PSA).
Each method measures the ES-performance (denoted by $\Per$) by its own means (see Sec.~\cref{sec:pcs}).
Lines 21-29 control which single PCS is used throughout an optimization.
The population is changed within the bounds $[\mu_\mathrm{min},\mu_\mathrm{max}]$ via a factor $\alpha_\mu>1$ using a simple schedule (Lines 31-37).
A generational idle (wait) time $\Delta_g>0$ can be set (Line 5).
Rescaling of the mutation strength is performed in Line 41 and will be discussed in Sec.~\cref{sec:csa_analysis}.
The algorithms are later tested on a set of test functions.
%The analysis of the CSA and PCS in the subsequent sections will be based on properties of the intermediate multi-recombinative $\muilam{\mu}{\lambda}$-ES, see Alg.~\cref{alg:alg1}, on the sphere function $f(R) = R^2$ in the limit of large population sizes.
%\input{alg_csa.tex}%
%

\section{The CSA-ES for Large Population Sizes}
\label{sec:csa_analysis}

\subsection{Scaling of Generation Number and  Mutation Strength}
\label{sec:gen_num_sig}

In this section, important theoretical and experimental results regarding the scaling behavior of the CSA-ES in the limit of large population sizes are shown on the sphere function $f(R) = R^2$, $R = ||\vb{y}||$, $\vb{y}\in\mathbb{R}^N$.
It will be shown that different standard CSA-implementations have different adaptation properties w.r.t.~the population size $\mu$ and dimensionality $N$.
These results will enable to derive a generation number scaling for the ES in the limit of large $\mu$ to achieve a given (relative) target.
The results will illustrate why large differences in the generation number are observed for the standard CSA-implementations.
These results will also be relevant for adaptive population control in Sec.~\cref{sec:pcs}, showing notable differences of the performance depending on the chosen CSA and its adaptation properties.

% New definitions
\algnewcommand\algorithmicswitch{\textbf{switch}}
\algnewcommand\algorithmiccase{\textbf{case}}
\algnewcommand\algorithmicassert{\texttt{assert}}
\algnewcommand\Assert[1]{\State \algorithmicassert(#1)}%
% New "environments"
\algdef{SE}[SWITCH]{Switch}{EndSwitch}[1]{\algorithmicswitch\ #1\ \algorithmicdo}{\algorithmicend\ \algorithmicswitch}%
\algdef{SE}[CASE]{Case}{EndCase}[1]{\algorithmiccase\ #1}{\algorithmicend\ \algorithmiccase}%
\algtext{EndSwitch}{\textbf{end switch}}%
\algtext*{EndCase}%

\newcommand\Algphase[1]{%
	\vspace*{-.7\baselineskip}\Statex\hspace*{\dimexpr-\algorithmicindent-2pt\relax}\rule{\columnwidth}{0.4pt}%
	\Statex\hspace*{-\algorithmicindent}\textbf{#1}%
	\vspace*{-.7\baselineskip}\Statex\hspace*{\dimexpr-\algorithmicindent-2pt\relax}\rule{\columnwidth}{0.4pt}%
}

\begin{algorithm}[t]
	\small
	\caption{Population Size Control via $(\mu/\mu_I, \lambda)$-CSA-ES}
	\label{alg:pcs_algos}
	\begin{algorithmic}[1]
		\State $g \gets 0$ 
		\State $\operatorname{initialize\_CSA}(\vb{y}^{(0)}, \vb{s}^{(0)}, \sigma^{(0)}, c_\sigma, d_\sigma, D, E_\chi)$ %\vb{y}^{(0)}, \sigma^{(0)}, \vb{s}^{(0)}, c_\sigma, d_\sigma, D
		\State $\operatorname{initialize\_PCS\_generic}(\mu^{(0)}, \mu_\mathrm{min},\mu_\mathrm{max}, \vt, \alpha_\mu, \Delta_g)$
		\State $\operatorname{initialize\_PCS\_specific}(\beta, L, f_\mathrm{rec}^{(0)}, f_\mathrm{med}^{(0)}, \gmcm{\vb{p}}{0}, \gmcc{\vb{p}}{0})$
		\State $w \gets \Delta_g$
		\Repeat
		\Algphase{Standard $(\mu/\mu_I, \lambda)$-CSA-ES}
		\For{$l = 1, ..., \lambda$} 
		\State $\tilde{\vb{z}}_l \gets [\mathcal{N}(0,1), ..., \mathcal{N}(0,1)]$
		\State $\tilde{\vb{y}}_l \gets \vb{y}^{(g)} + \sigma^{(g)}\tilde{\vb{z}}_l$ 
		\State $\tilde{f}_l \gets f(\vb{\tilde{y}}_l)$
		\EndFor
		\State	$(\tilde{f}_{1;\lambda}, ..., \tilde{f}_{m;\lambda}, ..., \tilde{f}_{\mu;\lambda} ) \gets \operatorname{sort}(\tilde{f}_1, ..., \tilde{f}_\lambda )$
		\State $\vb{y}^{(g+1)} \gets \frac{1}{\mu}\SUMM \tilde{\vb{y}}_{m;\lambda}$ 
		\State $\rec{\vb{z}}^{(g+1)} \gets \frac{1}{\mu}\SUMM \vb{z}_{m;\lambda}$
		\State $\vb{s}^{(g+1)} \gets (1-c_\sigma)\vb{s}^{(g)} + \sqrt{\mu^{(g)} c_\sigma(2-c_\sigma)}\rec{\vb{z}}^{(g+1)}$
		\State CSA~\cref{eq:new_han_v1}: $\sigma^{(g+1)} \gets
		%\begin{cases} 
			 \sigma^{(g)}\EXP{\frac{1}{D}\qty(||\vb{s}^{(g+1)}||/E_\chi-1)}$
		\State CSA~\cref{eq:neq_han_v2}:
			$\sigma^{(g+1)} \gets
			\sigma^{(g)}\EXP{\frac{c_\sigma}{d_\sigma}\qty(||\vb{s}^{(g+1)}||/E_\chi-1)}$
		%\end{cases}$
		\vspace{0.5em}
		\Algphase{Population Control Strategy (PCS)}
		\State $f_\mathrm{rec}^{(g+1)} \gets f(\vb{y}^{(g+1)})$
		\State $\fmed{g+1} \gets \operatorname{median}(\tilde{f}_{1;\lambda}, ..., \tilde{f}_{m;\lambda}, ..., \tilde{f}_{\mu;\lambda})$

		\State $g_0 \gets g-L+1$
		
		\If{PCS = APOP $\mathbf{and}$ $g_0\geq0$}
			\State $\Per \gets \mathrm{get\_apop}()$
		\ElsIf{PCS = pcCSA $\mathbf{and}$ $g_0\geq0$}
			\State $\Per \gets\mathrm{get\_pccsa}()$
		\ElsIf{PCS = PSA}
			\State $\Per \gets\mathrm{get\_psa}()$
		\Else
			\State $\Per \gets 0$
		\EndIf
		
		\If{$w=0$} 
			\If{$\Per<0$}
			\State $\mu^{(g+1)} \gets \ceil{\alpha_\mu \mu^{(g)}}$
			\ElsIf{$\Per>0$}
			\State $\mu^{(g+1)} \gets \floor{\mu^{(g)}/\alpha_\mu}$
			\Else
				\State $\mu^{(g+1)} \gets \mu^{(g)}$
			\EndIf

			\If{$\mu^{(g)}\neq\mu^{(g+1)}$} 
				\State $\mu^{(g+1)} \gets \min \qty(\max(\mu^{(g+1)}, \mu_\mathrm{min}), \mu_\mathrm{max})$ 
				\State $w \gets \Delta_g$ 
				\State $\sigma^{(g+1)} \gets \sigma^{(g)} r_\sigma(\mu^{(g)}, \mu^{(g+1)})$
				\State $\mu\gets\mu^{(g+1)},\quad \lambda \gets \mathrm{round}(\mu^{(g+1)}/\vt)$
				\State $c_\sigma  \gets  c_\sigma(\mu),\hspace{0.5em} D  \gets  D(\mu),\hspace{0.5em}  d_\sigma  \gets  d_\sigma(\mu)$ 
			\EndIf
		\Else 
			\State $w \gets w-1$
		\EndIf
		
		\State $g \gets g+1$
		\Until termination criterion
	\end{algorithmic}
\end{algorithm}%

%In this section, scaling properties for the generation number $G$ and isotropic mutation strength $\sigma$ are derived based on the sphere progress rate $\varphi$.
%The results are later used for adaptive population control.
The subsequent derivation of the generation number as a function of the progress rate was already given in \cite[Sec. 2.4]{Bey00b}.
The progress rate $\varphi$ is defined as the expected residual distance change between two generations $g$ and $g+1$ as
\begin{align}\begin{split}\label{sec:dyn_phidef}
		\varphi^{(g)} \coloneqq R^{(g)}-\EV{R^{(g+1)}}.
	\end{split}
\end{align}
%
%For simplicity we set $R^{(g+1)} = \EV{R^{(g+1)}}$.
Due to the scale-invariance of the sphere, one can define the normalized quantities (denoted by $^*$) as
\begin{align}\begin{split}\label{eq:sign} 
		\varphi^* = \varphi N / R,\quad \sigma^* = \sigma N/R.
\end{split}\end{align}
A properly working $\sigma$-adaptation, such as the CSA, attains constant $\sigma^* $ and $\varphi^*$ (in expectation) on the sphere, giving linear convergence order \cite{Bey00b}. 
%Using \cref{eq:sign}, expression \cref{sec:dyn_phidef} is rewritten as
%
%\begin{align}\begin{split}\label{sec:dyn_exp1}
%		R^{(g+1)}/R^{(g)} = 1 -\varphi^*(\sigma^*)/N.
%	\end{split}
%\end{align}
%
Assuming constant $\varphi^*$, one can
derive the generation number $G \coloneqq g-g_0$ for a given relative change $R^{(g)}/R^{(g_0)}$ under the assumption $\varphi^*/N\ll1$ as \cite[(2.105)]{Bey00b}
\begin{equation}\label{sec:dyn_phidyn_v2}
	G = (N/\varphi^*)\ln(R^{(g_0)}/R^{(g)}).
\end{equation}
%
%In our case, we will assume small $\varphi^*(\sigma^*)\gtrapprox0$ due to relatively slow adaptation of $\sigma^*$ (illustrated later).
The next step is to derive $\varphi^*$ for large populations.
One can start with the (normalized) progress rate of the sphere derived in \cite[(6.54)]{Bey00b} with progress coefficient $\CMULAM$
\begin{align}\begin{split}\label{eq:sph_Ndep}
\varphi^* &= \frac{\CMULAM\sigma^*(1 + \sigma^{*2}/2\mu N)}{\sqrt{1+\sigma^{*2}/\mu N}\sqrt{1+\sigma^{*2}/2N}} \\
&\qquad-N\qty(\sqrt{1+\sigma^{*2}/\mu N}-1) + O\qty(N^{-1/2}).
\end{split}\end{align}
\begin{figure}[t]
	\centering
	\begin{subfigure}{\columnwidth}
		\centering
		\includegraphics[width=\mywidth]{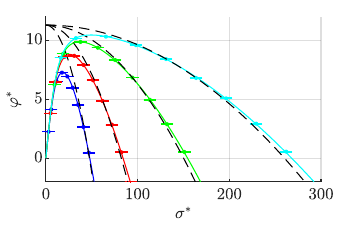}
		\includegraphics[width=\mywidth]{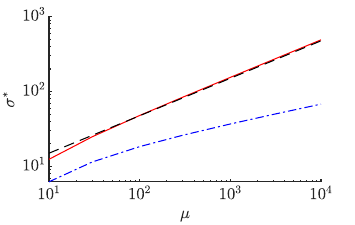}
		\caption{On the left, $\varphi^*(\sigma^*)$ at $N=100$ for 		$\mu=100,300,1000,3000$ and $\mu/\lambda=1/2$ are shown.
			The solid lines show \cref{eq:sph_Ndep} and the corresponding data points \cref{sec:dyn_phidef} averaged over $10^4$ trials and normalized using $\varphi^*=\varphi N/R$.
			The dashed line shows \cref{sec:dyn_phi_large}.
			On the right, \cref{eq:sph_Ndep} is used to numerically calculate the second zero $\sigma^*_0$ (red solid) and $\hat{\sigma}^* = \operatorname{arg\,max}\varphi^*(\sigma^*)$ (dash-dotted blue).
			The black dashed line shows approximation \cref{eq:signzero_approx}.}
		\label{fig:phi}
	\end{subfigure}
	\begin{subfigure}{\columnwidth}
		\centering
		\includegraphics[width=\mywidth]{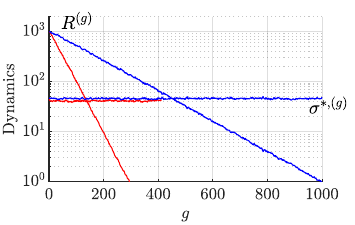}
		\includegraphics[width=\mywidth]{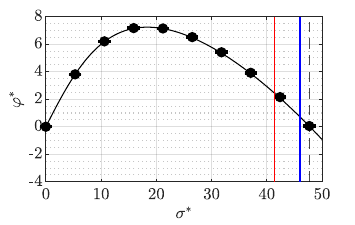}
		\caption{On the left, median dynamics of $\muilam{100}{200}$-CSA-ES for $N=100$ is shown (10 trials) using \cref{eq:sqrtN} (red) and \cref{eq:linN} (blue).
		%The converging curves show $R^{(g)}$ and the horizontal curves $\sigma^{*,(g)}$.
		On the right, $\varphi^*(\sigma^*)$ is shown (data points: simulated; black curve: \cref{eq:sph_Ndep}).
		Furthermore, the vertical lines mark measured median values from the left with $\signss\approx41.3$ (\cref{eq:sqrtN}, red, $\gamma\approx0.86$), $\signss\approx46.0$ (\cref{eq:linN}, blue, $\gamma\approx0.96$), and numerically obtained $\sigma^*_0\approx47.8$ (dashed black).}
		\label{fig:phi_dyn}
	\end{subfigure}
	\caption{Progress rate $\varphi^*$ and scale-invariant $\sigma^{*}$ on the sphere.}
	\label{fig:fig1}
\end{figure}\noindent
Progress rate \cref{eq:sph_Ndep} is visualized in Fig.~\cref{fig:fig1}.
Note that there is a range $\sigma^* \in (0,\signzero)$ where $\varphi^*>0$.
We will denote $\signzero$ as the analytic second zero (being an approximation to be derived), while $\sigma^*_0$ is the numerically obtained zero.
Expression~\cref{eq:sph_Ndep} needs to be simplified to enable closed-form solutions.
For large populations, one assumes $\mu N \gg \sigma^{*2}$ and $\CMULAM\simeq\CTHETA$ is only a function of the truncation ratio $\vartheta$ \cite[(6.113)]{Bey00b}.
Taylor-expansion in \cref{eq:sph_Ndep} yields $\sqrt{1+\sigma^{*2}/\mu N} = 1 + \sigma^{*2}/2\mu N + O( (\sigma^{*2}/\mu N)^2)$. 
Neglecting higher order terms, one gets a simplified expression
\begin{align}\begin{split}\label{eq:sph_med}
		\varphi^* &\simeq \CTHETA\sigma^*/\sqrt{1+\sigma^{*2}/2N} - \sigma^{*2}/2\mu.
\end{split}\end{align}
Intermediate result~\cref{eq:sph_med} needs to be approximated further.
As will be shown in Sec.~\cref{sec:csa}, the CSA-variants operate relatively slowly, i.e., closer to $\signzero$ than to $\sigma^*=0$.
Slow adaptation decreases $\sigma$ more slowly than fast adaptation while $R$ is decreased.
On the sphere, $\sigma^*$ remains constant, such that slow adaptation yields a higher $\sigma^*$-level via~\cref{eq:sign}.
Furthermore, large populations attain higher $\sigma^*$-values due to recombination.
Hence, we simplify \cref{eq:sph_med} by assuming $\sigma^{*2}/2N\gg1$, such that ``1" is neglected within the square-root.
One gets
\begin{equation}\label{sec:dyn_phi_large}
		\varphi^* \simeq \sqrt{2N}\CTHETA - \sigma^{*2}/2\mu.
\end{equation}
\mycomment{
At this point \cref{eq:signzero} can be further simplified.
It is assumed that a large population compared to $N$ is given, i.e., $\mu\gg\sqrt{N}$.
For optimization with population control under high multimodality, this assumption is reasonable as large population are needed to obtain high success rates.
One has
\begin{align}\begin{split}\label{eq:signzero_deriv}	
		\signzero &= \ParE{N\ParR{1 + \frac{8\CTHETA^2\mu^2}{N}}^{1/2}-N}^{1/2} \\
		&\simeq \ParE{N\ParR{\frac{8\CTHETA^2\mu^2}{N}}^{1/2}-N}^{1/2}\quad(\mu^2 \gg N) \\
		&= \ParE{(8N)^{1/2}\CTHETA\mu-N}^{1/2} \\
		&= (8N)^{1/4}(\CTHETA\mu)^{1/2}\ParE{1-\frac{N}{(8N)^{1/2}\CTHETA\mu}}^{1/2} \\
		&= (8N)^{1/4}(\CTHETA\mu)^{1/2}\ParE{1-\BigO{\frac{\sqrt{N}}{\mu}}} \quad(\text{Taylor expansion}, \mu\gg\sqrt{N}),
\end{split}\end{align}	
such that one obtains after neglecting $\BigO{\sqrt{N}/\mu}$ the result	
}
The zero of approximation \cref{sec:dyn_phi_large} is easily obtained as
\begin{align}\begin{split}\label{eq:signzero_approx}
		\signzero \simeq (8N)^{1/4}(\CTHETA\mu)^{1/2}.
\end{split}\end{align}
Figure~\cref{fig:phi} shows \cref{eq:sph_Ndep} and \cref{sec:dyn_phi_large} (left) and the second zero $\signzero$ (right).
%Note that the numerical solution of \cref{eq:sph_Ndep} will be denoted by $\sigma^*_0$.
One observes excellent agreement of \cref{eq:sph_Ndep} with one-generation experiments of \cref{sec:dyn_phidef}.
Furthermore, approximation \cref{sec:dyn_phi_large} improves with increasing $\mu$ (and $N$), which was expected from the underlying assumptions.
For \cref{eq:signzero_approx} one observes good agreement of $\signzero$ with $\sigma^*_0$ for increasing $\mu$.
In Fig.~\cref{fig:phi}, note that the optimal $\signopt$ decreases in relation to $\signzero$ for fixed $N$ with increasing $\mu$.
Using PCS on highly multimodal functions (with adequate global structure), small mutation strengths are undesirable since they increase the probability of local convergence.
Hence, the approach is to characterize the CSA-adaptation on the sphere in terms of a steady-state $\signss$ w.r.t.~the second zero $\signzero$ and not w.r.t.~the optimal value.
This will also be justified by the experiments of standard CSA implementations. 
Introducing a scaling factor $0<\gamma<1$ (slow-adaptation: $\gamma\lessapprox1$), one sets
\begin{align}\begin{split}\label{eq:signzero_approx_gam_v2}
		\signss = \gamma\signzero = \gamma(8N)^{1/4}(\CTHETA\mu)^{1/2}.
\end{split}\end{align}
Depending on the CSA parametrization (cumulation constant, damping), population size, and $N$, the ES achieves a steady-state $\signss$ w.r.t.~$\signzero$ according to a ratio $\gamma = \signss/\signzero$.
An example of measured values is shown in Fig.~\cref{fig:phi_dyn}.
Depending on the chosen CSA, one observes different measured $\gamma=\signss/\sigma^*_0$.
The limit $\gamma\rightarrow1$ corresponds to vanishing progress (stagnation) on the sphere.
The adaptation speed of the CSA will be characterized in terms of $\gamma$.
Assuming the CSA attains a constant $\gamma$, \cref{eq:signzero_approx_gam_v2} is inserted into \cref{sec:dyn_phi_large}, which is in turn used in \cref{sec:dyn_phidyn_v2}.
The progress rate yields
\begin{align}\begin{split}\label{eq:psa_ss_phinorm1}
		\varphi^* &\simeq \CTHETA(2N)^{1/2}(1-\gamma^2).
\end{split}\end{align}
After inserting \cref{eq:psa_ss_phinorm1} into \cref{sec:dyn_phidyn_v2}, one obtains for $G = g-g_0$
%
%\begin{mymathbox}
\begin{equation}\label{sec:dyn_phi_exp3}
	G \simeq \sqrt{N}\frac{\ln(R^{(g_0)}/R^{(g)})}{\sqrt{2}\CTHETA(1-\gamma^2)}.
\end{equation}
%\end{mymathbox}\noindent
%
For constant $\gamma$ and sufficiently large $\mu$, the generation number scales independent of $\mu$ with $\sqrt{N}$ (given a relative change of $R$).
This is a remarkable result which is tested later in experiments.
It is also useful when defining a time scale for the adaptation of PCS in Secs.~\cref{sec:schedule} and \cref{sec:pcs}.

Various PCS employ additional corrections of $\sigma$ after $\mu$ has been changed between two generations \cite{NA18, NH17}.
The idea is referred to as $\sigma$-rescaling and should accelerate the $\sigma$-adaptation further. 
However, this has effects on the stability of the CSA (see Sec.~\cref{sec:schedule}).
Given the results on the sphere in \cref{eq:signzero_approx_gam_v2} with $\sigma^* = \sigma N/R$, we derive a $\sqrt{\mu}$-law for the $\sigma$-rescaling $r_\sigma\coloneqq\sigma^{(g+1)}/\sigma^{(g)}$ (Alg.~\cref{alg:pcs_algos}, Line 41).
Since the distance to the optimizer does not change during $\sigma$-rescaling, one has $R^{(g+1)} = R^{(g)}$.
From \cref{eq:signzero_approx_gam_v2} one sees that $\sigma^{(g)} = \sigma_{\mathrm{ss}}^{*,(g)} R^{(g)}/N = K \sqrt{\mu^{(g)}} R^{(g)}/N $ and
$\sigma^{(g+1)} = \sigma_{\mathrm{ss}}^{*,(g+1)} R^{(g+1)}/N = K \sqrt{\mu^{(g+1)}} R^{(g+1)}/N$ (constant $K>0$). 
Thus, one gets $r_{\sigma}=\sqrt{\mu^{(g+1)}/\mu^{(g)}}$ and finally
%
%
%\begin{align}\begin{split}
%		\label{eq:sigma_rescale}
%		R^{(g+1)} &= R^{(g)} \\
%		\sigma^{(g+1)} &= \sigma^{(g)}\sigma^{*,(g+1)}/\sigma^{*,(g)},
%\end{split}\end{align}
%
%such that with $r_\sigma = \sigma^{*,(g+1)}/\sigma^{*,(g)} = \sigma^{(g+1)}/\sigma^{(g)}$ one gets
%
\begin{equation}
	\label{eq:sigma_rescale_sqrtMU}
	\sigma^{(g+1)} = \sigma^{(g)}\sqrt{\mu^{(g+1)}/\mu^{(g)}}.
\end{equation}
%\end{mymathbox}\noindent
%
Alternatively, in the case of $N\rightarrow\infty$ and $\mu \ll N$, \cref{eq:sph_Ndep} can be used to derive the well-known progress rate
\begin{align}\begin{split}\label{eq:sph_Ninf}
		\varphi^* &= \CMULAM\sigma^* - \sigma^{*2}/2\mu.
\end{split}\end{align}
In this case, the optimal value (maximizer) and the second zero of \cref{eq:sph_Ninf} are $\hat{\sigma}^*=\CMULAM\mu$ and $\signzero=2\CMULAM\mu$.
Based on these results, one obtains mutatis mutandis a linear scaling law for the $\sigma$-rescaling as
\begin{equation}
	\label{eq:sigma_rescale_linMU}
	\sigma^{(g+1)} = \sigma^{(g)}\mu^{(g+1)}/\mu^{(g)}.
\end{equation}
While \cref{eq:sigma_rescale_linMU} is not expected to yield better results (since $\mu\ll N$), it will still be tested as it was used in \cite{NA18} (see supplementary material \cref{ap:sigma_rescale} for a derivation on the sphere).
\subsection{Analysis of the CSA-ES}
In this section, the adaptation properties of the CSA-ES are studied.
The goal is to first investigate $\gamma = \gamma(\mu,N)$ from \cref{eq:signzero_approx_gam_v2} and later apply the results to obtain $G = G(\mu,N)$ from \cref{sec:dyn_phi_exp3}.
The CSA-ES has already been investigated in \cite{OB24b} for three standard CSA-parametrizations.
Hence, only the main results will be stated and more details are found in \cite{OB24b}.
%For the subsequent investigation, three commonly chosen implementations of the CSA-ES will be tested, see also Alg.~\cref{alg:alg1}.
The cumulation path $\vb{s}$ of the CSA-adaptation is given in terms of cumulation constant $\cs$ and recombined mutation direction $\rec{\vb{z}}$
\begin{align}\label{eq:csa_s}
	\gvec{s}{g+1} &= (1-\cs)\gvec{s}{g} + \sqrt{\cs(2-\cs)\mu} \rec{\vb{z}}^{(g+1)}. 
\end{align}
Then, the path length $\norm{\vb{s}^{(g+1)}}$ is measured and compared to its expected result under random selection.
The first update rule for the $\sigma$-change is chosen according to \cite{Han98} 
\begin{align}\begin{split}
		\label{eq:new_han_v1}
		\sigma^{(g+1)} = \sigma^{(g)}\EXP{\frac{1}{D}\qty(\norm{\vb{s}^{(g+1)}}/E_\chi-1)},
\end{split}\end{align}
where $D$ is a damping factor.
Alternatively, a slightly different update rule \cite[(44)]{hansen2023cma} with damping $d_\sigma$ yields
\begin{align}\begin{split}
		\label{eq:neq_han_v2}
		\sigma^{(g+1)} = \sigma^{(g)}\EXP{\frac{c_\sigma}{d_\sigma}\qty(\norm{\vb{s}^{(g+1)}}/E_\chi-1)},
\end{split}\end{align}
which is often chosen in recent implementations of the CSA.
$E_\chi$ is the expected value of a chi-distributed random variate $\chi \sim \norm{\mathcal{N}(\vb{0},\vb{1})}$.
One usually uses the approximation $E_\chi\simeq\sqrt{N}\qty(1-1/4N+1/21N^2)$ for large $N$.
%\begin{align}\begin{split}\label{eq:new_han_Echi} 			
%		E_\chi\simeq\sqrt{N}\qty(1-1/4N+1/21N^2).
%\end{split}\end{align} 
%
The CSA variants under investigation are parameterized as
\begin{subequations}
\begin{align}
		&\text{Eq.~\cref{eq:new_han_v1}}\quad \text{with} \quad c_\sigma = 1/\sqrt{N},\quad D = c_\sigma^{-1}. \label{eq:sqrtN} \\
		&\text{Eq. \cref{eq:new_han_v1}}\quad \text{with} \quad c_\sigma = 1/N,\quad D = c_\sigma^{-1} \label{eq:linN}\\
		\begin{split}
		&\text{Eq. \cref{eq:neq_han_v2}}\quad \text{with}\quad c_\sigma = \frac{\mu+2}{N+\mu+5},\qq{and} \\
		&d_\sigma = 1+c_\sigma+2\max\qty(0,\sqrt{(\mu-1)/(N+1)}-1). \label{eq:han}
	\end{split}
\end{align} 
\end{subequations}%
CSA implementations \cref{eq:sqrtN} and \cref{eq:linN} were investigated in more detail in \cite{HO01,Han98}.
\cite{Han98} derives the inverse proportionality $D = c_\sigma^{-1}$ with $c_\sigma = N^{-a}$ for $\frac{1}{2}\leq a \leq 1$ based on theoretical and experimental investigations on the sphere ($\mu\ll N$, see also \cref{eq:sph_Ninf}).
%Note that the result \cref{eq:sqrtN} will re-appear
%during the subsequent steady-state analysis for large populations.
CSA \cref{eq:han} is a newer implementation that is part of the default CMA-ES, see also \cite{hansen2023cma} ($\mu_\mathrm{eff}$ due to weighted recombination was replaced by $\mu$).

\mycomment{The subsequent analysis will show that the three CSA variants have distinct adaptation properties on the sphere as a function of $\mu$ and $N$.
Furthermore, $c_\sigma$ and $D$ from \cref{eq:sqrtN} are re-derived using a steady-state analysis on the sphere by assuming $\mu \gg N$.}

\begin{figure}[t]
	\centering
	\begin{subfigure}{0.48\columnwidth}
		\centering
		\includegraphics[width=\mywidth]{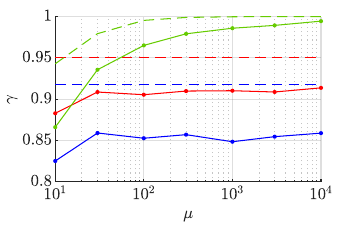}
		\caption{$N=10$.}
		\label{fig:compare_CSAs_a}
	\end{subfigure}
	\begin{subfigure}{0.48\columnwidth}
		\centering
		\includegraphics[width=\mywidth]{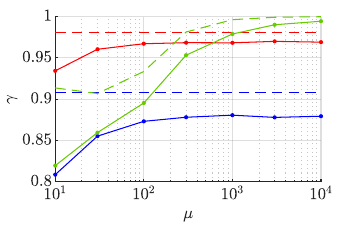}
		\caption{$N=100$}
		\label{fig:compare_CSAs_b}
	\end{subfigure}
	\begin{subfigure}{0.48\columnwidth}
		\centering
		\includegraphics[width=\mywidth]{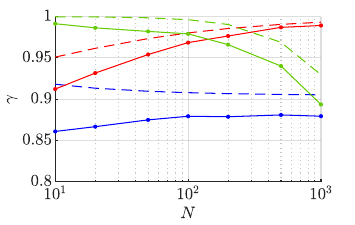}
		\caption{$\mu=1000$.}
		\label{fig:compare_CSAs_c}
	\end{subfigure}
	\begin{subfigure}{0.48\columnwidth}
		\centering
		\includegraphics[width=\mywidth]{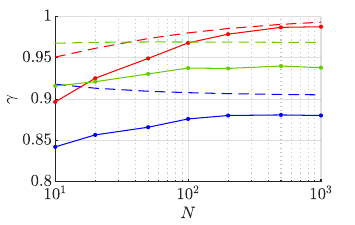}
		\caption{$\mu=2N$.}
		\label{fig:compare_CSAs_d}
	\end{subfigure}
	\caption{Steady-state ratio $\gamma$ on the sphere function for $\vt=1/2$. Measured ratio $\signss/\sigma^*_0$ (solid, with dots) compared to $\gamma$ from \cref{eq:csa_gamma_b_v2} (dashed) for the CSA variants \cref{eq:sqrtN} (blue), \cref{eq:linN} (red), and \cref{eq:han} (green).}
	\label{fig:compare_CSAs_gamma}
\end{figure}

As a first step, basic adaptation properties of the three variants are identified.
Note that $\cs$ and $D$ of \cref{eq:sqrtN} and \cref{eq:linN} are independent of $\mu$.
Changing the population size should not have large influence on the adaptation characteristics.
For \cref{eq:han}, $c_\sigma$ and $d_\sigma$ requires further analysis under the assumption $\mu\gg N$.
$c_\sigma$ yields simply
\begin{align}\begin{split}\label{eq:csigma_han}
		c_\sigma &= \frac{\mu(1+2/\mu)}{\mu(1+N/\mu+5/\mu)} \overset{\mu\rightarrow\infty}{\simeq} 1.
\end{split}\end{align}
The cumulation time parameter $c_\sigma$ approaches ``1" as $\mu$ is increased, which results in a faster cumulation in \cref{eq:csa_s}.
\mycomment{we introduce $d_\sigma = 1+c_\sigma+g(N,\mu)$ with $g(N,\mu) \coloneqq 2\max\qty(0,\sqrt{\frac{\mu-1}{N+1}}-1)$.
	Now, $d_\sigma$ is inserted into the exponential of \cref{eq:neq_han_v2}, which yields 
	%
	%\begin{align}\begin{split}\label{eq:neq_han_v2_check}
	\begin{math}		\EXP{\frac{c_\sigma\qty(\norm{\vb{s}^{(g+1)}}/E_\chi-1)}{1+c_\sigma+g(N,\mu)}} = \EXP{\frac{\norm{\vb{s}^{(g+1)}}/E_\chi-1}{1+1/c_\sigma+g(N,\mu)/c_\sigma}}.
	\end{math}	
	%\end{split}\end{align}
	%
	Comparing the last exponential with \cref{eq:new_han_v1}, one can derive the resulting damping parameter as $D = 1+1/c_\sigma+g(N,\mu)/c_\sigma$.
	Note that the proportionality $d_\sigma\propto c_\sigma^{-1}$ holds.
	Further analysis of the term $g(N,\mu)/c_\sigma$ yields for large $\mu \gg N$
	\begin{align}\begin{split}\label{eq:neq_han_v2_check3}
			&g(N,\mu)/c_\sigma = 2\max\qty(0,\sqrt{\frac{\mu-1}{N+1}}-1)\frac{N+\mu+5}{\mu+2} \\
			&\simeq 2\qty(\sqrt{\frac{\mu-1}{N+1}}-1)\frac{N+\mu+5}{\mu+2} \qq{(large $\mu$)} \\
			%&\simeq 2\sqrt{\frac{\mu-1}{N+1}}\frac{N+\mu+5}{\mu+2} \qq{(neglect 1 for large $\mu$)} \\
			%&= 2\sqrt{\frac{\mu(1-1/\mu)}{N+1}}\frac{1+N/\mu+5/\mu}{1+2/\mu} \qq{(factor out)} \\
			&\simeq 2\sqrt{\mu/N} \qq{($\mu\rightarrow\infty$)}.
	\end{split}\end{align}
	For the last line of \cref{eq:neq_han_v2_check3}, the ``$-1$" after the square root was neglected and $N+1\simeq N$ and $\mu-1\simeq \mu$ were applied.
}% mycomment
Now the factor $\cs/d_\sigma$ is brought into a similar form as $1/D$ in \cref{eq:new_han_v1}.
One finds in \cite{OB24b} that $c_\sigma/d_\sigma$ yields a damping factor $D = 1+1/c_\sigma+g(N,\mu)/c_\sigma$ with $g(N,\mu) \coloneqq 2\max\qty(0,\sqrt{\frac{\mu-1}{N+1}}-1)$.
Hence, the damping term $g(N,\mu)/c_\sigma$ yields the scaling
\begin{align}\begin{split}\label{eq:neq_han_v2_check3}
		g(N,\mu)/c_\sigma &\simeq 2\sqrt{\mu/N} \qq{($\mu\rightarrow\infty$).}
\end{split}\end{align}
The resulting damping $D$ of \cref{eq:han} scales with $\sqrt{\mu}$ according to \cref{eq:neq_han_v2_check3} for fixed $N$ and $\cs$ approaches one with \cref{eq:csigma_han}.
Hence, CSA \cref{eq:han} employs a $\mu$-dependent damping which is in contrast to \cref{eq:sqrtN} and \cref{eq:linN}.
If $\mu$ is changed during adaptive population control, the CSA adaptation characteristic in terms of $\gamma$ will be affected, which is shown later in Fig.~\cref{fig:compare_CSAs_gamma}.
%For the steady-state analysis of the CSA on the sphere, we will investigate the update equations \cref{eq:csa_s} and \cref{eq:new_han_v1}.
In \cite{OB24b} the three CSA-implementations are investigated and the CSA update equations are expressed in the sphere steady-state, where the scale-invariant mutation strength $\sigma^*$ and progress rate $\varphi^*$ are constant in expectation (see also Fig.~\cref{fig:phi_dyn}).
By applying certain approximations (assuming sufficiently slow progress $\varphi^*/N\ll1$ and large dimensionality $N$), this approach allows for closed-form of the CSA sphere steady-state as a function of the given cumulation constant and damping.
By including this dependency into a single constant $b$ \cite[(58)]{OB24b}
\begin{equation}\label{eq:csa_b_gamma}
	b \coloneqq \qty(\cs D/(1-\cs) + \sqrt{2}\CTHETA D/\sqrt{N})^{-1},
\end{equation}
one can derive the respective $\gamma$ from \cref{eq:signzero_approx_gam_v2} as a function of $b$ (assuming $1/\sqrt{2}<\gamma<1$) as \cite[(57)]{OB24b}
\begin{align}\begin{split}\label{eq:csa_gamma_b_v2}
		\gamma = \sqrt{\frac12\qty(\sqrt{1+b^2}-b + 1)}.
\end{split}\end{align}
The result \cref{eq:csa_gamma_b_v2} is interesting as it relates the CSA-parameters and $N$ to a scale factor $\gamma$ on the left side.
Demanding $\gamma$ to be constant and independent of $\mu$ and $N$, one must choose $D\propto\sqrt{N}$ and $\cs = D^{-1}$ within $b$ (this holds with $O(1/\sqrt{N})$).
This is only fulfilled by CSA~\cref{eq:sqrtN}.
In this case, the analytic solution of $\gamma \approx 0.90$ ($\mu/\lambda=1/2$) was derived as an approximation \cite[(63)]{OB24b}.
For \cref{eq:linN} one has $D \propto N$, and for \cref{eq:han} $D\propto\sqrt{\mu/N}$ via \cref{eq:neq_han_v2_check3}.
The limit $D\rightarrow\infty$ ($0<\cs<1$) yields $b\rightarrow0$ and $\gamma\rightarrow1$.
In this limit, $\signss\rightarrow\signzero$ in \cref{eq:signzero_approx_gam_v2} and the progress rate vanishes $\varphi^*\rightarrow0$ by approaching its second zero.
Hence, \cref{eq:linN} becomes increasingly slow for large $N$, while \cref{eq:han} becomes slower for increasing ratio $\sqrt{\mu/N}$.

% mycomment

In Fig.~\cref{fig:compare_CSAs_gamma}, experiments on the sphere are shown to compare the CSA-variants with the predicted adaptation behavior in terms of $\gamma$.
To this end, the dimensionality $N$ and population size $\mu$ are varied.
%The dimensionality is varied at fixed (large $\mu$) which is required by the underlying approximations.
%$\mu$ is varied for different constant $N$-values which mimics the case of having adaptive population control at given $N$-values.
The measured steady-state $\signss$ is averaged over at least 10 trials
and the median is taken over the measured $\sigma^{*,(g)}$ (the median is necessary due to a slightly skewed distribution of $\sigma^*$ at small $N$).
The measured $\signss$ is normalized by $\sigma^*_0$
(numerically obtained second zero of \cref{eq:sph_Ndep}\footnote{For $N\gtrapprox100$ one may numerically calculate $\sigma^*_0$ using \cref{eq:sph_Ndep}.
	For $N<100$, the progress rate zero is calculated from one-generation experiments of \cref{sec:dyn_phidef} to have a higher accuracy. 
	Slow adaptation yields $\sigma^*$ very close to $\sigma^*_0$ such that the accuracy of \cref{eq:sph_Ndep} is reduced due to missing $O\qty(N^{-1/2})$-terms.}), 
yieding a reference value for $\gamma$. 
In Figs.~\cref{fig:compare_CSAs_a} and 	\cref{fig:compare_CSAs_b}, the measured ratio remains relatively constant for sufficiently large $\mu$ for \cref{eq:sqrtN} and \cref{eq:linN}.
CSA \cref{eq:han} shows a significant increase of $\gamma$ as $\mu$ increases, which was expected from its damping $D$.
Deviations between the predicted $\gamma$ (dashed) and measurement (solid) are expected to occur due to the underlying (necessary) approximations applied in \cite{OB24b}.
%This is related to the approximations necessary to obtain a closed-form solution.
%Recall that $R^{(g+1)}/R^{(g)}=1$ was necessary to evaluate $\mathrm{E}[s_A^{(g+1)}]$ and higher order terms %in \cref{eq:csa_signs} were neglected (details in supplementary material \cref{ap:csa}).
In Fig.~\cref{fig:compare_CSAs_c}, the dimensionality is varied.
CSA \cref{eq:sqrtN} remains relatively constant, while $\gamma$ of \cref{eq:linN} increases for larger $N$, which was expected from its damping $D\propto N$.
CSA~\cref{eq:han} shows a decreasing $\gamma$ due to $D\propto \sqrt{\mu/N}$.
In Fig.~\cref{fig:compare_CSAs_d}, both $\mu$ and $N$ are varied together by maintaining $\mu=2N$.
CSA \cref{eq:sqrtN} and \cref{eq:han} remain approximately constant, while for \cref{eq:linN} $\gamma\rightarrow1$.
As expected, only \cref{eq:sqrtN} maintains an approximately constant ratio (best agreement for large $N$ and $\mu\gg N$) at $\gamma\lessapprox0.9$, which agrees satisfactory with the prediction \cref{eq:csa_gamma_b_v2}.
Furthermore, it realizes lower $\gamma$-levels, which leads to higher progress rates $\varphi^*$ (cf.~Fig.~\cref{fig:phi_dyn}).
Note that all three CSA yield relatively large $\gamma$-values between 0.8 and 1.
This means they achieve comparably low progress rates, which holds especially when compared to a self-adaptive ES ($\gamma \gtrapprox 0.6$, see \cite{OB24b}).
However, on highly multimodal problem instances, slower adaptation is beneficial to achieve higher success rates \cite{SB23,OB24}.

\begin{figure}[t]
	\centering
	\includegraphics[width=\mywidth]{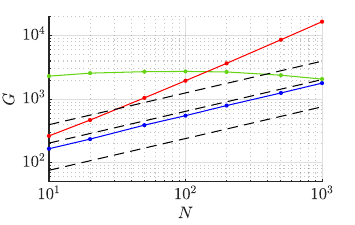}
	\includegraphics[width=\mywidth]{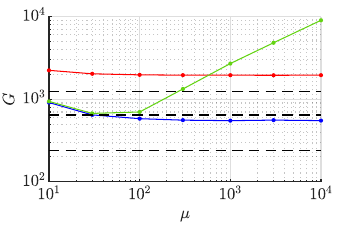}
	\caption{Number of generations $G$, see \cref{sec:dyn_phi_exp3}, to reach the target $R^{(g)}/R^{(g_0)}\!=\!10^{-6}$ using a $\muilam{1000}{2000}$-ES (left) and $N=100$ (right).
		The average over 10 runs is taken.
		The solid lines show \cref{eq:sqrtN} in blue, \cref{eq:linN} in red, and \cref{eq:han} in green.
		$G$ from \cref{sec:dyn_phi_exp3} at $\gamma=0.9$ is displayed in dash-dotted black. 
		The lower and upper dashed lines show $\gamma=0.8, 0.95$, respectively.
	}
	\label{fig:new_fig6}
\end{figure}

\mycomment{Note that all three CSA yield relatively large $\gamma$-values between 0.8 and 1.
This means they achieve comparably low progress rates according to Fig.~\cref{fig:phi_sign}.
Furthermore, their $\sigma^*$-levels are not optimal, i.e., they are not maximizing $\varphi^*$.
On highly multimodal functions with adequate global structure, slow-adaptation is usually beneficial to achieve high success rates.
However, it is not necessarily beneficial in terms of function evaluations (see Sec.~\cref{sec:pcs}).
It is interesting to note that an ES with mutative self-adaptation of $\sigma$ also maintains an approximately constant $\gamma\gtrapprox0.6$.
However, its adaptation at default values on the sphere is faster compared to the CSA.
This is illustrated in Sec.~\cref{ap:sa} in the supplementary material.}
Now that the results for $\gamma$ have been investigated in Fig.~\cref{fig:compare_CSAs_gamma}, one can study the generation number $G$ from \cref{sec:dyn_phi_exp3}.
Figure~\cref{fig:new_fig6} shows $G$ evaluated at constant $\mu$ (left) and constant $N$ (right).
CSA~\cref{eq:sqrtN} yields very good agreement with  \cref{sec:dyn_phi_exp3} due to the relatively constant $\gamma$.
On the left, one observes a $\sqrt{N}$-law for $G$.
On the right, $G$ remains asymptotically constant for increasing $\mu$.
CSA~\cref{eq:linN} shows a faster increase of $G$ than $\sqrt{N}$ (left), which can be attributed to the increased damping ($\gamma\rightarrow1$) at large $N$.
On the right, it also remains asymptotically constant for large $\mu$, requiring more generations due to slower adaptation.
CSA~\cref{eq:han} shows a different characteristic, being comparably slow at high ratios $\mu/N$ and fast at low ratios mostly due to its damping $D$.
The $G$-asymptotic for $\mu\rightarrow\infty$ is a notable result.
From a modeling perspective, one could argue that CSA~\cref{eq:sqrtN} shows important (desired) properties for PCS on sphere-like functions.
The adaptation is comparably fast and one observes a generational speedup of $G$ as $\mu$ is increased from small initial values.
At large $\mu$, asymptotic behavior is observed.
CSA~\cref{eq:han} slows down its adaptation as $\mu$ is increased, which can be detrimental in terms of efficiency (see later discussion of Fig.~\cref{fig:ras}).

	\section{Population Control Schedule}
\label{sec:schedule}
\begin{figure}[t]
	\begin{subfigure}{1\columnwidth}
		\centering
		\includegraphics[width=\mywidth]{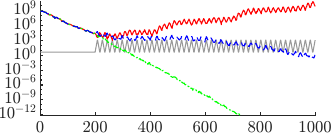} 
		\includegraphics[width=\mywidth]{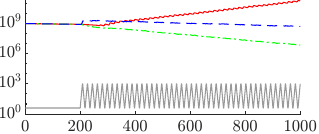} 
		\includegraphics[width=\mywidth]{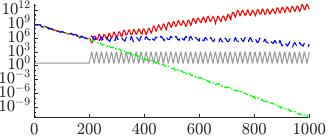}
		\includegraphics[width=\mywidth]{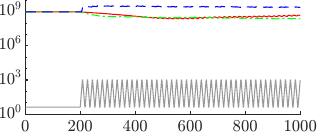}
		\includegraphics[width=\mywidth]{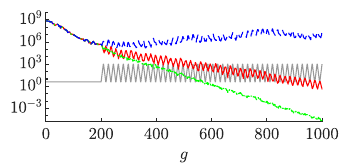}
		\includegraphics[width=\mywidth]{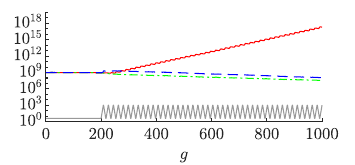}
		\caption{No waiting time $\Delta_g = 0$ (left: $N=10$, right: $N=1000$).}
		\label{fig:schedule}
	\end{subfigure}
	\begin{subfigure}{1\columnwidth}
		\centering
		\includegraphics[width=\mywidth]{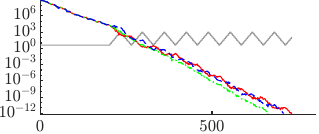} 
		\includegraphics[width=\mywidth]{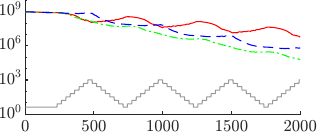} 
		\includegraphics[width=\mywidth]{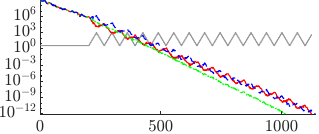}
		\includegraphics[width=\mywidth]{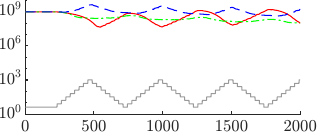}
		\includegraphics[width=\mywidth]{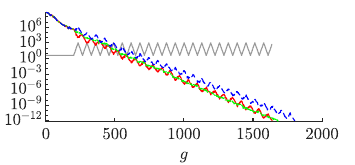}
		\includegraphics[width=\mywidth]{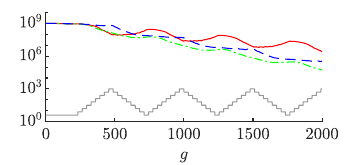}
		\caption{Waiting time $\Delta_g = \ceil{\sqrt{N}}$ (left: $N=10$, right: $N=1000$).}
		\label{fig:schedule_wait}
	\end{subfigure}
	\caption{Stability of CSA on the sphere using predefined $\mu^{(g)}$-schedule.
	The gray lines show $\mu^{(g)}$ and the colored lines the $R^{(g)}$-dynamics of \cref{eq:r1} (red), \cref{eq:r2} (green), and \cref{eq:r3} (blue). 
	The tested CSAs are \cref{eq:sqrtN}, \cref{eq:linN}, and \cref{eq:han}, from top to bottom, respectively for (a) and (b).}
	\label{fig:schedules}
\end{figure}

Before adaptive population control is investigated further, the CSA-stability w.r.t.\ changes of $\mu$ is investigated.
The PCS will later introduce additional method-specific feedback in the ES-dynamics.
Having effects of $\sigma$- and $\mu$-adaptation combined makes the analysis more difficult.
Hence, we introduce a simple $\mu^{(g)}$-schedule on the sphere function, generating artificial $\mu$-changes.
The schedule is defined as being constant within the first 200 generations.
Then, $\mu$ is increased using a factor $\alpha_\mu$ as $\mu^{(g+1)} = \ceil{\mu^{(g)}\alpha_\mu}$ until $\mu_\mathrm{max}$ is reached.
After reaching $\mu_\mathrm{max}$, it is decreased as $\mu^{(g+1)} = \floor{\mu^{(g)}/\alpha_\mu}$ until $\mu^{(0)}$ is reached.
This oscillation is repeated multiple times to check the stability.
Furthermore, the effect of rescaling $\sigma$ (via $\sigma^{(g+1)} = r_\sigma\sigma^{(g)}$) on the stability of the CSA is tested.
Given results \cref{eq:sigma_rescale_sqrtMU} and \cref{eq:sigma_rescale_linMU}, one can summarize the three tested rescaling methods as%
\begin{subequations}\begin{align}
r_\sigma &= 1\qq{(no rescaling)} \label{eq:r1} \\
r_\sigma &= (\mu^{(g+1)}/\mu^{(g)})^{1/2} \label{eq:r2} \\
r_\sigma &= \mu^{(g+1)}/\mu^{(g)}. \label{eq:r3}
\end{align}\end{subequations}
The experiments in Figs.~\cref{fig:schedule} and \cref{fig:schedule_wait} show the convergence dynamics on the sphere function.
Two different values for $N$ are tested based on Fig.~\cref{fig:compare_CSAs_gamma}, where large differences for $\gamma(N)$ could be observed.
In both experiments, $\alpha_\mu=2$ is chosen with $\mu^{(0)}\!=\!\mu_\mathrm{min}\!=\!4$ and $\mu_\mathrm{max}\!=\!1024$.
The dynamics are initialized at $\vb{s}^{(0)}=\vb{1}$ and $\sigma^{(0)} = \signzero R/N$ using \cref{eq:signzero_approx}, which reduces undesired initialization effects.
In Fig.~\cref{fig:schedule_wait}, a waiting time $\Delta_g = \ceil{\sqrt{N}}$ is applied after a population change based on the result $G\propto\sqrt{N}$ from \cref{sec:dyn_phi_exp3}.
This introduces more time for the CSA to adapt to the $\mu$-change.
Note that no waiting is implemented in Fig.~\cref{fig:schedule}.

In Fig.~\cref{fig:schedule}, one observes that no rescaling ($r_\sigma = 1$) leads to instabilities (divergence) of the CSA in most cases, while $r_\sigma = (\mu^{(g+1)}/\mu^{(g)})^{1/2}$ shows very good results in terms of convergence speed for all configurations.
It also shows the least amount of $R^{(g)}$-oscillations along its convergence.
Scaling $r_\sigma = \mu^{(g+1)}/\mu^{(g)}$ shows mixed results.
It helps to stabilize the CSA in some cases.
However, it would be more suitable for $\mu$-control with $\mu\ll N$ due to progress rate \cref{eq:sph_Ninf}.
The instabilities are related to the CSA cumulation constant $\cs$, the damping $D$, and the applied $r_\sigma$.
Due to the high rate of change of $\mu$, the CSA is constantly adapting to the changing population.
In the worst case, one observes divergence due to $\sigma^*$ being outside the positive progress range $\sigma^*\in(0,\signzero)$, see Fig.~\cref{fig:phi_dyn}.
Note that CSA~\cref{eq:sqrtN} (first row) shows consistent results as $N$ is increased.
On the other hand, CSA~\cref{eq:han} converges with $r_\sigma=1$ due to the higher damping at small $N=10$ (although with large oscillations of $R$), but diverges due to lower damping at large $N=1000$.
In Fig.~\cref{fig:schedule_wait}, the waiting time $\Delta_g = \ceil{\sqrt{N}}$ yields more time for the CSA to adapt to the current population size $\mu^{(g)}$.
Again, \cref{eq:r2} shows the most stable results.
As expected, $\Delta_g >0$ stabilizes all the CSA variants and for all $r_\sigma$-choices.
The obtained results show that \cref{eq:r2} is the preferred rescaling of $\sigma$ on the sphere in terms of stability.
In Sec.~\cref{sec:pcs} it will be shown that certain PCS are more sensitive to changes of $\sigma$ than others.
The influence of $r_\sigma$ with active PCS will also be studied.

	\section{Adaptive Population Control Strategies}
\label{sec:pcs}

For the subsequent investigations, three state-of-the-art PCS are implemented based on \cite{NH17} (APOP), \cite{HB16} (pcCMSA), and \cite{NA18} (PSA).
These algorithms were designed and tested for adaptive population control on ES for a broader range of functions (noisy and/or multimodal, details below).
As for the analysis, the idea is to implement simplified versions of the PCS.
Hence, only the performance measuring routines are implemented.
Note that the population control subroutines of Alg.~\cref{alg:pcs_algos} are given in Alg.~\cref{alg:pcs_subr}, including the pseudocode for APOP, pcCSA, and PSA.
%The actual change of $\mu$ is done using the simple schedule $\ceil{\mu^{(g)}\alpha_\mu}$ and $\floor{\mu^{(g)}/\alpha_\mu}$ for increase and decrease, respectively.
A simple routine for the change of $\mu$ is realized according to Alg.~\cref{alg:pcs_algos} ({Lines 31-37}).
We omit the introduction of additional damping parameters for the $\mu^{(g)}$-change (as done in \cite{NA18, NH17}, analogous to $D$ in \cref{eq:new_han_v1}) since it adds more complexity.
Furthermore, \cref{eq:r2} is used for the $\sigma$-rescaling on all methods.

The APOP was introduced in \cite{NH17} for the CMA-ES and tested on a larger set of noiseless functions.
The basic idea is to count the number of (median-)fitness deteriorations within the last $L$ generations (in \cite{NH17} they use a fixed length $L=5$).
Defining the median over selected $f$-values as $\fmed{g} \coloneqq \operatorname{median}(f_{m;\lambda}^{(g)})$, $m=1,...,\mu$, the difference is given by
\begin{align}
	\label{eq:apop_exp1}
	\Delta f^{(g)} = \fmed{g}-\fmed{g-1}.
\end{align}
Then, the ratio $P_f$ is evaluated by counting the occurrence of $\Delta f^{(g)}>0$.
One has $L-1$ differences for $L$ values of $\fmed{g}$.
Using the indicator function $\mathbbm{1}$, the ratio yields
\begin{align}
	\label{eq:apop_exp2}
	P_f = \frac{1}{L-1} \sum_{i=0}^{L-2} \mathbbm{1}[\Delta f^{(g-i)} > 0].
\end{align}
A threshold $\Thr_f = 1/5$ is chosen in \cite{NH17} based on empirical studies and will be adopted.
$P_f > \Thr_f$ triggers a population increase due to insufficient performance, for $P_f = \Thr_f$ no change occurs, and for $P_f < \Thr_f$ the population is decreased.
As for all the PCS considered here, the respective threshold value $\Thr$ has significant influence on the performance of the PCS and exchanging the CSA will lead to notable performance differences on simple test functions.

The pcCMSA was introduced in \cite{HB16} and further analyzed in \cite{BH17}.
It was originally introduced as a covariance-matrix self-adaptive ES for population control on noisy functions and later tested on multimodal functions in \cite{NA18}.
In this paper, only the hypothesis test for convergence is implemented using a CSA, calling it pcCSA.
In \cite{HB16} it is argued that stagnation or divergence behavior coincides with a non-negative trend within the observed
fitness value dynamics of the ES (for minimization). 
For a trend analysis, a regression model of the parental recombinant fitness sequence $f(\vb{y}^{(g)})$ of length $L$ is used and a hypothesis test on the slope is done.
The (fluctuating) fitness dynamics $f$ is modeled assuming a linear model with slope $a$, intercept $b$, and random normal fluctuations $\epsilon_i$ as $f^{(g)} = a g + b + \epsilon_i$.
Denoting the standard error of the estimated slope $\hat{a}$ as $s_{\hat{a}}$, the (standardized) test statistic is a $t$-distributed variate with $L-2$ degrees of freedom
\begin{align}
	\label{eq:pccmsa_exp7}
	T_{L-2} \sim (\hat{a}-a)/s_{\hat{a}}.
\end{align}
The hypothesis test is defined as $H_0$: $a\geq0$, indicating no significant trend and insufficient performance. 
The alternative $H_1$: $a<0$ indicates sufficient performance.
We will evaluate the $P$-value of the test ($P_{t,L-2}$ denoting the distribution function of $T$ with $L-2$ degrees of freedom)
\begin{align}
	\label{eq:pccmsa_Pval}
	P_H \coloneq P_{t,L-2}\qty(\hat{a}/s_{\hat{a}}),
\end{align}
rejecting $H_0$ at a significance level $P_H < 0.05$ ($\Thr_H=0.05$).
In this case, $\mu$ is decreased, or otherwise increased.

\begin{algorithm}[t]
	\small
	\caption{Population Control Strategies (PCS)}
	\label{alg:pcs_subr}
	\begin{algorithmic}[1]
		\State Measure performance $\Per$ in Alg.~\cref{alg:pcs_algos} by evaluation of: 
		
		\Algphase{$\mathbf{get\_apop}()$}
		\State $\vb{d} \gets \mathrm{diff}(\fmed{g_0:g})$
		\State $P_f \gets \mathrm{sum}(\vb{d}>0)/(L-1)$
		\State $\Per \gets \mathrm{perf}(P_f<\Thr_f$: $1$;\quad
		$P_f>\Thr_f$: $\!-1$;\quad
		$P_f=\Thr_f$: $0)$
		
		\Algphase{$\mathbf{get\_pccsa}()$}
		\State $\vb{g} \gets [g_0:g], \quad		\vb{f} \gets [f_\mathrm{rec}^{(g_0:g)}]$
		\State $\bar{g} \gets \mathrm{mean}(\vb{g}),\quad \bar{f} \gets \mathrm{mean}(\vb{f})$
		\State $\hat{a} \gets \frac{\sum_{i=1}^L(g_i-\bar{g}) (f_i-\bar{f})}{\sum_{i=1}^L(g_i-\bar{g})^2},\quad \hat{b} \gets \bar{f} - \hat{a}\bar{g}$	
		\State$ s_{\hat{a}} \gets \sqrt{\frac{\sum_{i=1}^L(f_i-\bar{f})^2}{(L-2)\sum_{i=1}^L(g_i-\bar{g})^2}}$	
		\State $t \gets \hat{a}/s_{\hat{a}}$
		\State $P_H \gets \mathrm{tcdf}(t,L-2)$ \Comment{distribution function of $T_{L-2}$}
		\State $\Per \gets \mathrm{perf}(P_H<\Thr_H$: $1$;
		\quad$P_H>\Thr_H$:\quad$\!-1$;\quad$P_H=\Thr_H$: $0)$
		
		\Algphase{$\mathbf{get\_psa}()$}
		\State $E_F \gets N/\mu^{(g)}$
		\State $\wt{\Delta}_m \gets \rec{\vb{z}}^{(g+1)}$
		\State $\wt{\Delta}_c \gets \frac{1}{\sqrt{2}}((\sigma^{(g+1)}/\sigma^{(g)})^2-1)\vb{1}$
		\State	$\gmcm{\vb{p}}{g+1} \gets (1-\beta)\gmcm{\vb{p}}{g} + \sqrt{\beta(2-\beta)/E_F}
		\widetilde{\Delta}_m$
		\State $\gmcc{\vb{p}}{g+1} \gets (1-\beta)\gmcc{\vb{p}}{g} + \sqrt{\beta(2-\beta)/E_F}\widetilde{\Delta}_c$
		\State $||\gmc{\vb{p}}{g+1}||^2 \gets ||\gmcm{\vb{p}}{g+1}||^2 + ||\gmcc{\vb{p}}{g+1}||^2$
		\State $\Per \gets \mathrm{perf(}$$\ppt\!<\!\Thr_\theta$: $-1$;
		\quad$\ppt\!>\!\Thr_\theta$: $1$; 
		\quad$\ppt\!=\!\Thr_\theta$: $0)$
		%\State $\Per \gets \mathrm{perf}$
		%($||\gmc{\vb{p}}{g+1}||^2<\Thr_\theta$: $-1$;
		%\quad
		%$||\gmc{\vb{p}}{g+1}||^2>\Thr_\theta$: $1$;
		%\quad
		%$||\gmc{\vb{p}}{g+1}||^2=\Thr_\theta$: $0$)
		
	\end{algorithmic}
\end{algorithm}\noindent

The core ideas of the PSA-CMA-ES were introduced in \cite{NA16} and further extended in \cite{NA18}.
It is based on the idea of measuring the change of search space parameters, namely the covariance matrix $\vb{C}$ and the mean search vector $\vb{y}$.
By accumulating the information using two cumulation paths, their respective lengths are used as an indicator of ES performance.
The key element is to distinguish random selection (insufficient performance) from non-random selection (sufficient performance).
The measured path lengths are used to change the population size, given a certain threshold.
In contrast to pcCSA and APOP, the performance is measured solely in search space and not in $f$-space 
\footnote{Note that \cite{NA18} and \cite{NA16} elaborate the similarities between their update equations and the natural gradient interpretation of the CMA-ES.
This is omitted at this point since we will investigate a simplification of the PSA-CMA-ES by only including the CSA without covariance matrix adaptation.}.%
The main idea is to define two cumulation paths for the change of the mean $\dm^{(g+1)} = \vb{y}^{(g+1)} - \vb{y}^{(g)}$, and the change of $\vb{C}$ and $\sigma$ as $\dS^{(g+1)} = (\sigma^{(g+1)})^2\vb{C}^{(g+1)} - (\sigma^{(g)})^2\vb{C}^{(g)}$. 
Then, $\dm^{(g+1)}$ and $\dS^{(g+1)}$ are transformed using their respective Fisher transformation matrices to achieve invariance w.r.t.~the chosen (normal) search space distribution.
Using a CSA-ES, i.e., $\vb{C}=\vb{I}$ (identity matrix), the update equations simplify significantly.
The detailed derivation is presented in the supplementary material \cref{ap:psa_analysis}.
The corresponding cumulation paths from \cref{ap:csa_pm} and \cref{ap:psa_sigmachange} yield
\begin{align}
		\gmcm{\vb{p}}{g+1} &= (1-\beta)\gmcm{\vb{p}}{g} + \sqrt{\beta(2-\beta)\mu/N} \rec{\vb{z}}^{(g+1)}.  \label{eq:csa_pm} \\
		\gmcc{\vb{p}}{g+1} &= (1-\beta)\gmcc{\vb{p}}{g} + \sqrt{\frac{\beta(2-\beta)\mu}{(2N)}}\qty[\frac{(\sigma^{(g+1)})^2}{(\sigma^{(g)})^2}-1]\vb{1}, \label{eq:psa_sigmachange}
\end{align}
with $\vb{1}=[1,\dots,1]$ of length $N$.
The PSA measures the squared norm of \cref{eq:csa_pm} and \cref{eq:psa_sigmachange}.
Aggregating both vectors into a single update vector $\vb{p}_\theta^{(g+1)} = (\gmcm{\vb{p}}{g+1}, \gmcc{\vb{p}}{g+1})$, one evaluates
\begin{equation}\label{eq:psa_components}
	\ppt = \ppm + \ppc.
\end{equation}
The results of \cref{eq:csa_pm} and \cref{eq:psa_sigmachange} are notable.
While \cref{eq:csa_pm} mirrors the cumulation of the CSA (using a different constant $\beta$ and normalization w.r.t.~$N$, see also CSA-update in \cref{eq:csa_s}), cumulation \cref{eq:psa_sigmachange} aggregates relative $\sigma$-changes.
The PSA works by measuring $\ppt=\ppm+\ppc$ and comparing it to the threshold $\Thr_\theta = 1.4$, cf.~Fig.~\cref{fig:psa_dyn}.
Under random selection, one observes $\ppm\approx1$ and $\ppc\gtrapprox0$, which indicates insufficient performance ($\ppt < \Thr_\theta$) and increases $\mu$.
Random selection with $\ppm=1$ can be obtained by evaluating $||\gmcm{\vb{p}}{g+1}||^2$ in \cref{eq:csa_pm}.
Assuming a steady-state in expectation, one has $\mathrm{E}[||\gmcm{\vb{p}}{g+1}||^2]=\mathrm{E}[||\gmcm{\vb{p}}{g}||^2]=||\vb{p}_m||^2$, $\mathrm{E}[\vb{p}_m^{(g)} \rec{\vb{z}}^{(g+1)}] = 0$ (no preferred search direction) and $\mathrm{E}[\norm{\rec{\vb{z}}}^2] = N/\mu$ from \cite[(5.2)]{Arn02},
such that one simply gets $||\vb{p}_m||^2=1$.
The analogous calculation holds for the steady-state $\mathrm{E}[||\vb{s}||^2] = N$ of the CSA under random selection.
From the update rules \cref{eq:new_han_v1} and \cref{eq:neq_han_v2} one infers that $\sigma$ does not change in expectation. Hence, the contribution of \cref{eq:psa_sigmachange} vanishes, giving $\ppc=0$.
If the selection is not random, both $\ppm$ and $\ppc$ yield relevant contributions and the population is controlled to maintain $\ppt\approx\Thr_\theta$.
Analytic investigations of $\ppm$ and $\ppc$ in the sphere steady-state are given in the supplementary material \cref{ap:psa_stst}.

\mycomment{
	Similar to the steady-state analysis of the CSA in \cite{OB24b}, one can look at steady-state properties of $\ppt$.
	This can be done by evaluating $||\vb{p}^{(g+1)}||^2$, setting $||\vb{p}^{(g+1)}||^2=||\vb{p}^{(g)}||^2=||\vb{p}||^2$ and taking the respective expected value.
	Evaluating $\EV{\ppm}$ under random selection, one has $\EV{\vb{p}_m^{T} \rec{\vb{z}}} = 0$ (no preferred search direction).
	The analogous observation holds for the mixed term of $\EV{||\vb{s}||^2}$ for the CSA.
	Hence, one has
	Evaluating $\ppm$ under random selection, one has $\EV{\vb{p}_m^{T} \rec{\vb{z}}} = 0$ (no preferred search direction) and $\EV{\norm{\rec{\vb{z}}}^2} = N/\mu$ \cite[(5.2)]{Arn02}
	The analogous observation holds for the mixed term of $||\vb{s}||^2$ for the CSA.
	%Dropping the mixed term, the resulting cumulation yields for large $g\rightarrow\infty$ the results $\EV{\ppm}\rightarrow1$ (and $||\vb{s}||^2 \rightarrow N$).
	Furthermore, $\sigma$ does not change under random selection (see \cref{eq:new_han_v1} and \cref{eq:neq_han_v2}) and $\ppc\rightarrow0$ vanishes.
	The PSA-CSA works by measuring $\ppt$ and comparing it to the threshold $\Thr_\theta = 1.4$.
	Under random selection, one has $\ppm\approx1$ and $\ppc\approx0$, which indicates insufficient performance ($\ppt < 1.4$) and increases the population.
	If the selection is not random, both $\ppm$ and $\ppc$ yield relevant contributions and the population is controlled to maintain $\ppt\approx1.4$.
}%
%It will be illustrated later that this control has advantages and disadvantages.
%
\begin{figure}[t]
		\centering
		\includegraphics[width=8.8cm]{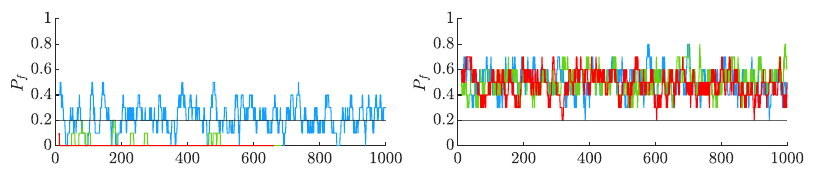} 
		\includegraphics[width=8.8cm]{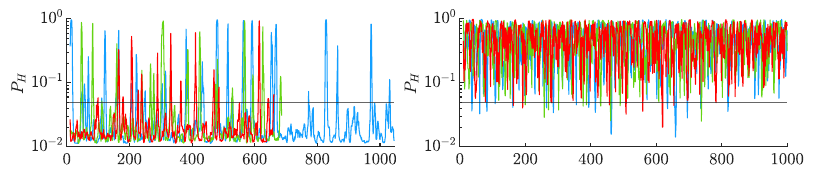} 
		\includegraphics[width=8.8cm]{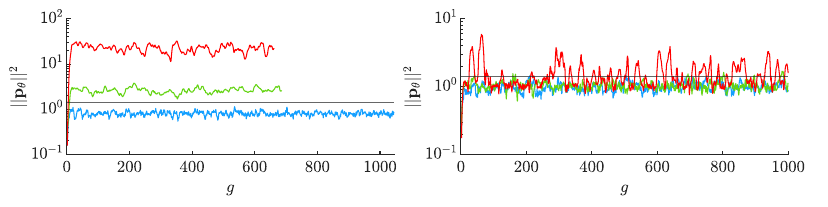} 
		\caption{Sphere (left column) and random function (right column) with CSA~\cref{eq:sqrtN} for $N=100$ using APOP (top, $L=10$), pcCSA (center, $L=10$), and PSA (bottom $\beta=1/10$). 
		The performance is measured with deactivated population control at $\mu=10,100,1000$ (blue, green, and red signals, respectively)
		The threshold $\Thr$ is shown in solid black.}
	\label{fig:new_pcs_deactPop}
\end{figure}
Before continuing the analysis, the performance measures of the PCS are evaluated with \emph{deactivated} population control.
In Fig.~\cref{fig:new_pcs_deactPop}, the sphere \cref{eq:fsph} and random function \cref{eq:fran} are evaluated at $N=100$ for three constant $\mu$-values.
The respective quantities $P_f$, $P_H$, and $\ppt$ are measured.
We are interested in the signal levels in relation to their thresholds.
%While fluctuations can be observed, we focus on the observed mean levels.
$P_f$ from APOP decreases for increasing $\mu$ on $f_\mathrm{sph}$.
Larger populations yield less deterioration of the median fitness due to a more robust search.
Note that at $\mu=10$, $P_f>\Thr_f=0.2$ most of the time.
With active population control, this would trigger an (undesired) $\mu$-increase on the sphere.
On $f_\mathrm{ran}$, the signals fluctuate around $P_f\approx0.5$ (above $\Thr_f$), successfully indicating bad performance due to random selection.
For the pcCSA, $P_H$ mostly stays below the significance level $\Thr_H=0.05$ on the sphere, which indicates good performance.
On $f_\mathrm{ran}$, $P_H$ lies mostly above the threshold, correctly indicating bad performance.
Sporadic $f$-fluctuations may (falsely) indicate good performance.
For the PSA, $\ppt$ shows different levels depending on $\mu$.
At $\mu=10$, $\ppt$ is smaller than the threshold $\Thr_\theta=1.4$. 
At $\mu=100,1000$, it lies above  it.
Similar to the APOP, the PSA will increase $\mu$ on the sphere function to reach $\ppt\approx1.4$.
This effect is due to the aggregation of relative $\sigma$-changes via \cref{eq:psa_sigmachange} which increase for larger $\mu$ due to faster convergence (large $\ppc$-contributions compared to $\ppm$, see supplementary material \cref{ap:psa_stst}).
On the random function, the three signals lie in the vicinity of $\ppt\approx1$ mostly due to $\ppm\approx1$ and vanishing $\ppc\gtrapprox0$ (fluctuations can be observed).
In general, the PSA detects random selection well.
%It can be explained noting that $\sigma^{(g+1)}/\sigma^{(g)} = R^{(g+1)}/R^{(g)}$ via \cref{eq:sph_med} with $\sigma^{(g)} = \sigma^{*,(g)}  R^{(g)} / N$ and assuming $\sigma^{*,(g)}=\sigma^{*,(g+1)}$.

\subsection{Method Comparison}
\label{sec:sph_ran}

For the comparison of the introduced PCS, two sets of parameters will be chosen according to Tab.~\cref{tab:configs} by the following argumentation.
The goal is to align the time scales of the PCS by which the performance of the ES is evaluated.
Both pcCSA and APOP aggregate $f$-values over a length of $L$ generations.
Using the derived time scale \cref{sec:dyn_phi_exp3}, we set $L=\ceil{N^{1/2}}$ for P1.
Similarly, the backward time horizon of the cumulation path scales as $1/\beta$, see \cite{hansen2023cma}.
To align the PSA with the array length $L$, we choose $\beta=1/N^{1/2}$.
Note that the CSA \cref{eq:sqrtN} also operates at $c_\sigma=1/N^{1/2}$.
For P1 one chooses a small $\alpha_\mu=1.05$, no waiting $\Delta_g$, and $r_\sigma$ \cref{eq:r2} according to Sec.~\cref{sec:schedule} to reduce the stress on the $\sigma$-adaptation ($\alpha_\mu=1.05 \ll 2$).
%As will be shown during the experiments, each PCS will have certain issues with P1 (also depending on the underlying CSA).
An alternative to P1 is the parameter set P2, including waiting times and performing larger $\mu$-changes.
This should stabilize the CSA additionally, see Fig.~\cref{fig:schedule_wait}.
Since $G\propto\sqrt{N}$ scales weakly with $N$, one may choose a constant value $L=10$ (and $\beta=1/10$).
As an example, $N=10$ would yield $L=\ceil{\sqrt{N}}=4$ and for $N=1000$ one has $L=32$.
Both results are not too far away from $L=10$.
A constant $L$ and $\beta$ schedule was also applied in \cite{NH17} and \cite{NA18}, however, large $N$ were not investigated.
Both sets require a comparable number of generations to reach $\mu_\mathrm{max}=1024$ starting from $\mu_\mathrm{min}=4$, with P1 yielding $g=89$ and P2 $g=78$ (if persistent $\mu$-increase is triggered).
\begin{table}[t]
	\centering
	\begin{tabular}{c|ccccc}
		& L & $\beta$ & $\alpha_\mu$ & $\Delta_g$ & $r_\sigma$ \\ \hline
		P1 & $\ceil{N^{1/2}}$ & $1/N^{1/2}$ & 1.05 & 0 & $(\mu^{(g+1)}/\mu^{(g)})^{1/2}$ \\ 
		P2 & $10$ & $1/10$ & 2 & 10 & $(\mu^{(g+1)}/\mu^{(g)})^{1/2}$ \\ 
	\end{tabular}
	\caption{Two parameter sets P1 and P2 for comparing the adaptive PCS.}
	\label{tab:configs}
\end{table}
\begin{table}[h!]
	\centering\tiny
	\begin{tabular}{ccccc}
		& $\mu_\text{25}$ & $\mu_\text{med}$ & $\mu_\text{75}$ & $F_t$ \\
		\hline
		S10 & 14 & 20 & 37 & 1.1e+04 \\
		\hline
		S100 & 23 & 95 & 318 & 3.3e+05 \\
		\hline
		S1000 & 29 & 130 & 523 & 1.9e+06 \\
		\hline
		N10 & 1024 & 1024 & 1024 & 1.9e+06 \\
		\hline
		N100 & 1024 & 1024 & 1024 & 1.9e+06 \\
		\hline
		N1000 & 1024 & 1024 & 1024 & 1.8e+06 \\
		\hline\hline
		S10 & 129.5 & 1024 & 1024 & 2.7e+05 \\
		\hline
		S100 & 4 & 4 & 7 & 1.8e+04 \\
		\hline
		S1000 & 4 & 4 & 4 & 1.6e+05 \\
		\hline
		N10 & 1024 & 1024 & 1024 & 1.9e+06 \\
		\hline
		N100 & 1024 & 1024 & 1024 & 1.9e+06 \\
		\hline
		N1000 & 1024 & 1024 & 1024 & 1.8e+06 \\
		\hline\hline
		S10 & 13 & 17 & 20 & 6.6e+03 \\
		\hline
		S100 & 33 & 41 & 56 & 6.7e+04 \\
		\hline
		S1000 & 97 & 126 & 163 & 7.4e+05 \\
		\hline
		N10 & 29 & 37 & 56 & 8.6e+04 \\
		\hline
		N100 & 343 & 594 & 927 & 1.2e+06 \\
		\hline
		N1000 & 1024 & 1024 & 1024 & 1.9e+06 \\
		\hline
	\end{tabular}
	\begin{tabular}{ccccc}
		& $\mu_\text{25}$ & $\mu_\text{med}$ & $\mu_\text{75}$ & $F_t$ \\
		\hline
		S10 & 8 & 16 & 32 & 1.2e+04 \\
		\hline
		S100 & 16 & 16 & 32 & 4.5e+04 \\
		\hline
		S1000 & 16 & 16 & 32 & 2.0e+05 \\
		\hline
		N10 & 1024 & 1024 & 1024 & 1.9e+06 \\
		\hline
		N100 & 1024 & 1024 & 1024 & 1.9e+06 \\
		\hline
		N1000 & 1024 & 1024 & 1024 & 1.9e+06 \\
		\hline\hline
		S10 & 4 & 4 & 8 & 3.7e+03 \\
		\hline
		S100 & 4 & 4 & 8 & 1.8e+04 \\
		\hline
		S1000 & 4 & 4 & 8 & 1.7e+05 \\
		\hline
		N10 & 1024 & 1024 & 1024 & 1.8e+06 \\
		\hline
		N100 & 1024 & 1024 & 1024 & 1.8e+06 \\
		\hline
		N1000 & 1024 & 1024 & 1024 & 1.7e+06 \\
		\hline\hline
		S10 & 8 & 8 & 16 & 5.0e+03 \\
		\hline
		S100 & 32 & 32 & 64 & 6.9e+04 \\
		\hline
		S1000 & 256 & 256 & 512 & 1.4e+06 \\
		\hline
		N10 & 16 & 32 & 64 & 1.1e+05 \\
		\hline
		N100 & 256 & 512 & 1024 & 1.2e+06 \\
		\hline
		N1000 & 1024 & 1024 & 1024 & 1.9e+06 \\
		\hline
	\end{tabular}
	\caption{CSA~\cref{eq:sqrtN} with APOP, pcCSA, PSA (top to bottom, respectively) with P1 and P2 (left and right column).}
	\label{tab:sqrtN}
\end{table}

\begin{table}[h!]
	\centering\tiny
	\begin{tabular}{ccccc}
		& $\mu_\text{25}$ & $\mu_\text{med}$ & $\mu_\text{75}$ & $F_t$ \\
		\hline
		S10 & 29 & 40 & 53 & 2.5e+04 \\
		\hline
		S100 & 38 & 137 & 429 & 1.3e+06 \\
		\hline
		S1000 & 37 & 177 & 761 & 1.5e+07 \\
		\hline
		N10 & 1024 & 1024 & 1024 & 1.9e+06 \\
		\hline
		N100 & 1024 & 1024 & 1024 & 1.9e+06 \\
		\hline
		N1000 & 1024 & 1024 & 1024 & 1.8e+06 \\
		\hline\hline
		S10 & 383 & 1024 & 1024 & 4.4e+05 \\
		\hline
		S100 & 593 & 800 & 975 & 3.2e+06 \\
		\hline
		S1000 & 11 & 18 & 31 & 9.4e+05 \\
		\hline
		N10 & 1024 & 1024 & 1024 & 1.9e+06 \\
		\hline
		N100 & 1024 & 1024 & 1024 & 1.9e+06 \\
		\hline
		N1000 & 1024 & 1024 & 1024 & 1.8e+06 \\
		\hline\hline
		S10 & 29 & 33 & 37 & 1.9e+04 \\
		\hline
		S100 & 429 & 474 & 550 & 2.1e+06 \\
		\hline
		S1000 & 1024 & 1024 & 1024 & 4.1e+07 \\
		\hline
		N10 & 133 & 201 & 287 & 4.3e+05 \\
		\hline
		N100 & 1024 & 1024 & 1024 & 1.9e+06 \\
		\hline
		N1000 & 1024 & 1024 & 1024 & 1.9e+06 \\
		\hline
	\end{tabular}
	\begin{tabular}{ccccc}
		& $\mu_\text{25}$ & $\mu_\text{med}$ & $\mu_\text{75}$ & $F_t$ \\
		\hline
		S10 & 32 & 32 & 64 & 2.9e+04 \\
		\hline
		S100 & 128 & 128 & 256 & 9.7e+05 \\
		\hline
		S1000 & 512 & 1024 & 1024 & 3.0e+07 \\
		\hline
		N10 & 1024 & 1024 & 1024 & 1.9e+06 \\
		\hline
		N100 & 1024 & 1024 & 1024 & 1.9e+06 \\
		\hline
		N1000 & 1024 & 1024 & 1024 & 1.9e+06 \\
		\hline\hline
		S10 & 4 & 8 & 16 & 7.6e+03 \\
		\hline
		S100 & 64 & 256 & 512 & 1.8e+06 \\
		\hline
		S1000 & 256 & 512 & 1024 & 2.3e+07 \\
		\hline
		N10 & 1024 & 1024 & 1024 & 1.8e+06 \\
		\hline
		N100 & 1024 & 1024 & 1024 & 1.8e+06 \\
		\hline
		N1000 & 1024 & 1024 & 1024 & 1.7e+06 \\
		\hline\hline
		S10 & 8 & 16 & 16 & 1.0e+04 \\
		\hline
		S100 & 256 & 512 & 512 & 2.4e+06 \\
		\hline
		S1000 & 1024 & 1024 & 1024 & 4.1e+07 \\
		\hline
		N10 & 64 & 128 & 256 & 4.0e+05 \\
		\hline
		N100 & 1024 & 1024 & 1024 & 1.9e+06 \\
		\hline
		N1000 & 1024 & 1024 & 1024 & 1.9e+06 \\
		\hline
	\end{tabular}
	\caption{CSA~\cref{eq:linN} with APOP, pcCSA, PSA (top to bottom, respectively) with P1 and P2 (left and right column).}
	\label{tab:linN}
\end{table}

\begin{table}[h!]
	\centering\tiny
	\begin{tabular}{ccccc}
		& $\mu_\text{25}$ & $\mu_\text{med}$ & $\mu_\text{75}$ & $F_t$ \\
		\hline
		S10 & 438 & 689 & 975 & 2.3e+06 \\
		\hline
		S100 & 97 & 399 & 1024 & 1.6e+06 \\
		\hline
		S1000 & 32 & 147 & 594 & 2.0e+06 \\
		\hline
		N10 & 1024 & 1024 & 1024 & 1.9e+06 \\
		\hline
		N100 & 1024 & 1024 & 1024 & 1.9e+06 \\
		\hline
		N1000 & 1024 & 1024 & 1024 & 1.8e+06 \\
		\hline\hline
		S10 & 1024 & 1024 & 1024 & 4.3e+06 \\
		\hline
		S100 & 4 & 6 & 9 & 2.0e+04 \\
		\hline
		S1000 & 4 & 4 & 4 & 1.8e+05 \\
		\hline
		N10 & 1024 & 1024 & 1024 & 1.9e+06 \\
		\hline
		N100 & 1024 & 1024 & 1024 & 1.9e+06 \\
		\hline
		N1000 & 1024 & 1024 & 1024 & 1.8e+06 \\
		\hline\hline
		S10 & 928 & 1024 & 1024 & 4.1e+06 \\
		\hline
		S100 & 40.5 & 59 & 103 & 2.8e+05 \\
		\hline
		S1000 & 88 & 109 & 135 & 5.7e+05 \\
		\hline
		N10 & 740 & 974 & 1024 & 1.5e+06 \\
		\hline
		N100 & 761 & 1024 & 1024 & 1.7e+06 \\
		\hline
		N1000 & 1023 & 1024 & 1024 & 1.8e+06 \\
		\hline
	\end{tabular}
	\begin{tabular}{ccccc}
		& $\mu_\text{25}$ & $\mu_\text{med}$ & $\mu_\text{75}$ & $F_t$ \\
		\hline
		S10 & 1024 & 1024 & 1024 & 3.5e+06 \\
		\hline
		S100 & 128 & 512 & 1024 & 2.1e+06 \\
		\hline
		S1000 & 16 & 32 & 32 & 2.3e+05 \\
		\hline
		N10 & 1024 & 1024 & 1024 & 1.9e+06 \\
		\hline
		N100 & 1024 & 1024 & 1024 & 1.9e+06 \\
		\hline
		N1000 & 1024 & 1024 & 1024 & 1.9e+06 \\
		\hline\hline
		S10 & 512 & 1024 & 1024 & 3.1e+06 \\
		\hline
		S100 & 4 & 4 & 8 & 1.9e+04 \\
		\hline
		S1000 & 4 & 4 & 8 & 1.8e+05 \\
		\hline
		N10 & 1024 & 1024 & 1024 & 1.8e+06 \\
		\hline
		N100 & 1024 & 1024 & 1024 & 1.8e+06 \\
		\hline
		N1000 & 1024 & 1024 & 1024 & 1.7e+06 \\
		\hline\hline
		S10 & 16 & 16 & 32 & 2.4e+04 \\
		\hline
		S100 & 32 & 64 & 512 & 6.2e+05 \\
		\hline
		S1000 & 128 & 256 & 256 & 1.3e+06 \\
		\hline
		N10 & 128 & 512 & 1024 & 1.1e+06 \\
		\hline
		N100 & 512 & 1024 & 1024 & 1.4e+06 \\
		\hline
		N1000 & 512 & 1024 & 1024 & 1.6e+06 \\
		\hline
	\end{tabular}
	\caption{CSA~\cref{eq:han} with APOP, pcCSA, PSA (top to bottom, respectively) with P1 and P2 (left and right column).}
	\label{tab:han}
\end{table}

The test functions to be investigated are chosen as follows.
The goal is to have a simple benchmark set to test basic, but essential properties of the PCS together with the underlying CSA.
The properties are performance on a unimodal function, behavior under high noise, and performance on a highly multimodal function, all at varying dimensionality.
To this end, we choose
\begin{subequations}
	\begin{align}
		f_\mathrm{sph}(\vb{y}) &\coloneqq \textstyle \sum_i^{N} y_i^2 \label{eq:fsph}\\
		f_\mathrm{ran}(\vb{y}) &\coloneqq \mathcal{N}(0,1) \label{eq:fran} \\
		f_\mathrm{ras}(\vb{y}) &\coloneqq \textstyle\sum_i^{N}[y_i^2 + A(1-\cos(\alpha y_i))]. \label{eq:fras}
	\end{align}
\end{subequations}
On unimodal functions, the population size should be kept small to reduce the number of function evaluations.
On the random function, it should reach its maximum value and be kept large.
The reason is that large populations reduce the expected residual distance for a sphere under high noise (see \cite{AB00d}).
On the Rastrigin function, it should be dynamically changed since local attraction is only relevant within a certain range \cite{SB23}.
For  both large and small $R^2=\sum_i y_i^2$, $f_\mathrm{ras}$ behaves like a quadratic function.

\begin{figure}[t]
	\centering
	\includegraphics[width=8.8cm]{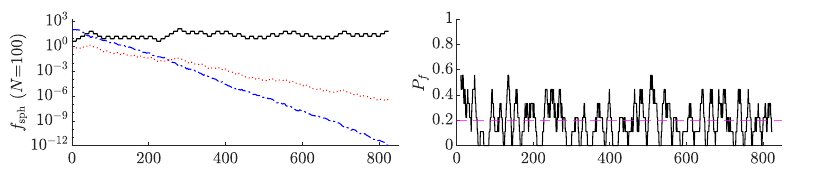} 
	\includegraphics[width=8.8cm]{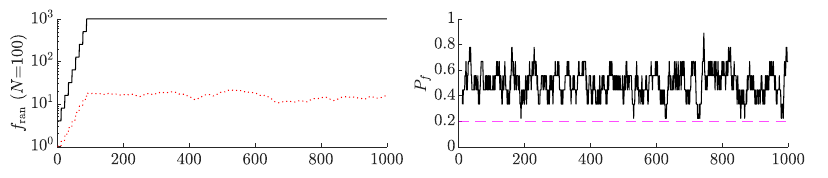} 
	\includegraphics[width=8.8cm]{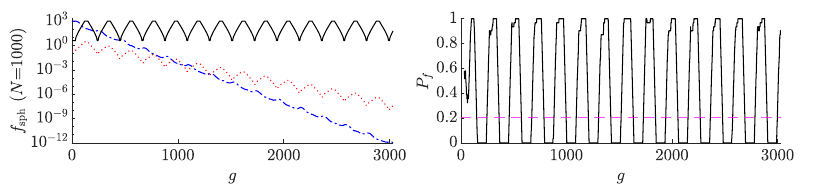} 
	\caption{APOP-dynamics with CSA~\cref{eq:sqrtN} for S100 (P2) at the top, N100 (P2) in the center, and undesired $\mu$-oscillations for S1000 (P1) at the bottom.}
	\label{fig:apop_dyn}
\end{figure}

\begin{figure}[t]
	\centering
	\includegraphics[width=8.8cm]{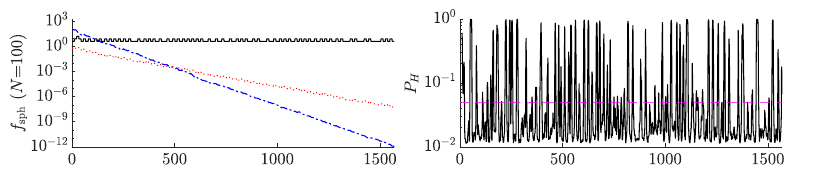} 
	\includegraphics[width=8.8cm]{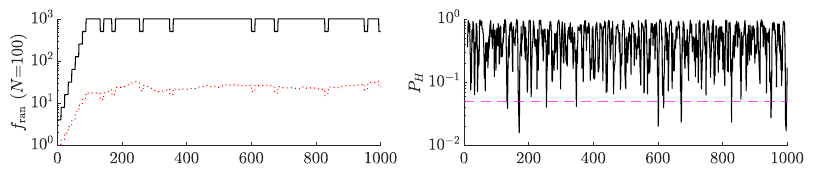} 
	\includegraphics[width=8.8cm]{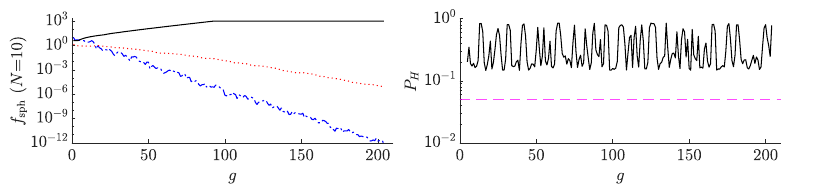} 
	\caption{pcCSA-dynamics with CSA~\cref{eq:sqrtN} for S100 (P2) at the top, N100 (P2) in the center, and undesired $\mu$-increase for S10 (P1) at the bottom.}
	\label{fig:pccsa_dyn}
\end{figure}

\begin{figure}[t]
	\centering
	\includegraphics[width=8.8cm]{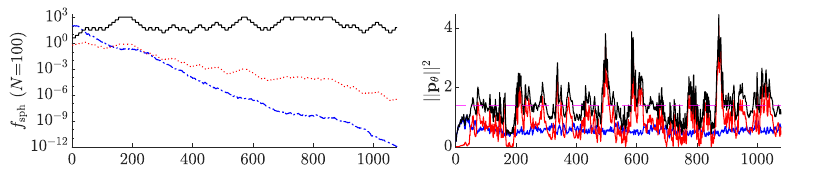} 
	\includegraphics[width=8.8cm]{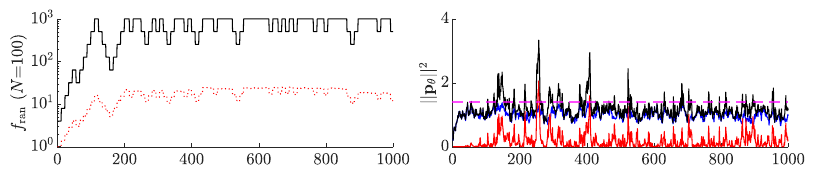} 
	\includegraphics[width=8.8cm]{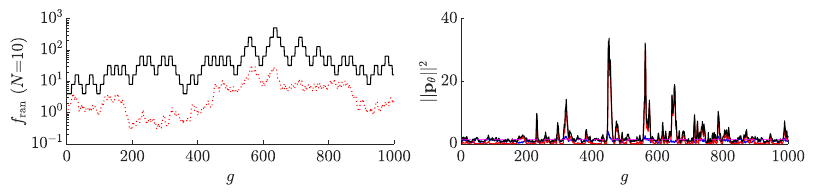} 
	\caption{PSA-dynamics with CSA~\cref{eq:han} for S100 (P2) at the top, N100 (P2) in the center, and insufficient $\mu$-level for N10 with CSA~\cref{eq:sqrtN} at the bottom.
		On the right, one has $\ppt$ (black), $\ppm$ (blue), and $\ppc$ (red).
	}
	\label{fig:psa_dyn}
\end{figure}

A series of experiments on $f_\mathrm{sph}$ (denoted by ``S") and $f_\mathrm{ran}$ (``N" for noise) is shown in Tab.~\cref{tab:sqrtN}, \cref{tab:linN}, and \cref{tab:han}.
The number denotes the dimensionality (e.g.~S10 is $f_\mathrm{sph}$ at $N=10$).
$\mu_{25}$, $\mu_\mathrm{med}$, and $\mu_{75}$ measure the 25-th, 50-th, and 75-th percentile of the occurred $\mu^{(g)}$.
They are an indicator of the overall $\mu$-level and distribution width.
$F_r$ measures the number of function evaluations.
Exemplary dynamics from the tables are shown in Figs.~\cref{fig:apop_dyn}, \cref{fig:pccsa_dyn}, and \cref{fig:psa_dyn}.
In general, one observes a high variation of the performance on $f_\mathrm{sph}$ and $f_\mathrm{ran}$, depending on the CSA.
The best configuration for APOP and pcCSA is shown in Tab.~\cref{tab:sqrtN} (CSA~\cref{eq:sqrtN}) for P2 (see top right and center right).
They maintain low $\mu$-levels on the $f_\mathrm{sph}$ and high, stable $\mu$-levels on $f_\mathrm{ran}$ (examples in Figs.~\cref{fig:apop_dyn} and \cref{fig:pccsa_dyn}).
The PSA shows mixed results (bottom right).
The main issue is insufficient performance (low $\mu$) for N10.
This is critical, e.g., when optimizing multimodal functions at small $N$ where high $\mu$-levels are needed for global convergence.
The overall best configuration for the PSA uses CSA~\cref{eq:han} (for which it was designed in \cite{NA18}), see Tab.~\cref{tab:han} (P2, bottom right).
However, it shows somewhat elevated $\mu$-levels on $f_\mathrm{sph}$ (also reported by \cite{NA18}).


Now the examples of Figs.~\cref{fig:apop_dyn}, 	\cref{fig:pccsa_dyn}, and	\cref{fig:psa_dyn} are discussed in more detail.
In the left plots, one has $\mu^{(g)}$ (solid black), 
$f^{(g)}$ (dash-dotted blue) and $\sigma^{(g)}$ (dotted red).
On the right, the corresponding performance measures are shown in black with $P_f$ (APOP), $P_H$ (pcCSA), and $\ppt$ (PSA, $\ppm$ in blue and $\ppc$ in red). 
The threshold values are shown as dashed magenta.
The respective top and center plots show examples where the PCS works relatively well.
The bottom plot shows an example where undesired behavior was observed.
Note that for $f_\mathrm{ran}$ no $f$-values are shown (random noise).
The APOP in Fig.~\cref{fig:apop_dyn} on $f_\mathrm{sph}$ controls $\mu$, such that $P_f$ is close to the threshold 0.2.
On $f_\mathrm{ran}$, $P_f\approx0.5$ lies above the threshold due to selection on a random function.
Hence, $\mu$ is kept at the maximum value.
In the bottom plot, zero waiting $\Delta_g = 0$ (via P1) introduces unnecessary oscillations of $\mu$.
They are related to constant changes of the $f$-distribution, such that \cref{eq:apop_exp1} falsely detects fitness deterioration.
Waiting $\Delta_g > 0$ is necessary to remove these issues (used in P2).
An example of the changing $f$-distribution is shown in the supplementary material \cref{ap:apop_add}.
The pcCSA in Fig.~\cref{fig:pccsa_dyn} shows very low $\mu$-levels on $f_\mathrm{sph}$ which is desired.
On $f_\mathrm{ran}$ the hypothesis test detects stagnation reliably and keeps $\mu$ large.
In the bottom plot, the small data set $L=\ceil{N}=4$ yields large standard errors of \cref{eq:pccmsa_Pval}, falsely indicating bad performance and increasing $\mu$ (example shown in the supplementary material \cref{ap:pccsa_add}).
The pcCSA requires a larger sample size $L$ to overcome this issue (see P2 with $L=10$).
The PSA in Fig.~\cref{fig:psa_dyn} measures
$\ppt$ (black) via $\ppm$ (blue) and $\ppc$ (red).
On $f_\mathrm{sph}$, $\ppm$ and $\ppc$ have similar contributions.
On $f_\mathrm{ran}$, $\ppm\approx1$ and $\ppc\gtrapprox0$, indicating random selection (see discussion below \cref{eq:psa_sigmachange}).
However, fluctuations of $\sigma$ may generate spurious larger $\ppc$-contributions, yielding $\ppt$ above the threshold and reducing $\mu$.
The bottom example shows issues of the PSA to increase $\mu$ on $f_\mathrm{ran}$ due to the mentioned fluctuations.
This occurs for CSA~\cref{eq:sqrtN}, but is less pronounced for CSA~\cref{eq:han} since the latter adapts $\sigma$ significantly slower due to $D\propto\sqrt{\mu/N}$ from \cref{eq:neq_han_v2_check3}.
The PSA requires slow adaptation for small $N$ to correctly detect random selection.
Note that this effect is not related to the applied $r_\sigma=\rsigsqrt$, but to the underlying CSA (see supplementary material \cref{ap:psa_add}).

One major observation can be made from Tab.~\cref{tab:linN} and \cref{tab:han}.
Recall that both CSAs \cref{eq:linN} and \cref{eq:han} show increasingly slow adaptation within respective limits, see ratio approaching ``1" in Fig.~\cref{fig:compare_CSAs_c}.
Very slow adaptation can significantly deteriorate the PCS performance on $f_\mathrm{sph}$.
In Tab.~\cref{tab:linN}, larger $N$ show large $\mu_\mathrm{med}$-levels and in Tab.~\cref{tab:han}, large $\mu_\mathrm{med}$-levels occur for small $N$.
In general, bad performance at slow adaptation on $f_\mathrm{sph}$ can be explained in terms of measured $f$-values using Fig.~\cref{fig:phi_dyn}.
Since the adaptation is very slow, the ES operates at very large $\sigma^*$ close to $\signzero$.
It progresses very slowly and selects worse fitness values more often.
The deterioration is reflected within the pcCSA by the hypothesis test mostly indicating stagnation.
Similarly, the APOP counts significantly more worse fitness changes via 	\cref{eq:apop_exp2}, leading to unnecessary $\mu$-increase on $f_\mathrm{sph}$.
%Note that the results could be improved (to some extent) by choosing larger adaptation times $L\propto N$ and $\beta\propto N^{-1/2}$.
The PSA shows mixed results.
In Tab.~\cref{tab:linN}, it works well for S10 (small $\mu_\mathrm{med}$), but yields $\mu_\mathrm{med}=\mu_\mathrm{max}$ for S1000.
It requires a different CSA, i.e., adaptation speed as a function of $\mu$ and $N$, to achieve better performance (see Tab.~\cref{tab:han}).

Comparing the experiments of parameter sets P1 and P2 in Tab.~\cref{tab:sqrtN}, \cref{tab:linN}, and \cref{tab:han} shows mixed results.
There are cases in which one option is either better or worse than the other and the results depend on both CSA and PCS.
However, P2 with $\Delta_g>0$ is preferred for APOP due to stability (see discussion of Fig.~\cref{fig:apop_dyn}).
Furthermore, P2 yields improved results for pcCSA (hypothesis test for small $N$) and in general good results for PSA.
Hence, the parameter set P2 will be used as default in Sec.~\cref{sec:comp_ras} for further investigations.

\subsection{Comparison on the Rastrigin Function}
\label{sec:comp_ras}

Now that basic properties of the PCS- and CSA-variants have been studied on $f_\mathrm{sph}$ and $f_\mathrm{ran}$, a simplified benchmark is performed on the Rastrigin function $f_\mathrm{ras}$.
The reason for choosing Rastrigin is that sufficiently large populations are needed to find the global attractor \cite{SB23,OB24}.
However, local attraction has only a limited range, such that adaptive PCS should keep the population size small far away from the global attractor (global quadratic structure) and within a local (or global) attractor.
The goal is to investigate the performance of the best configurations from Sec.~\cref{sec:sph_ran}.
Furthermore, the PCS will be compared against the CSA-variants with constant $\mu=\mumax$.

The tested parameter sets of Rastrigin are chosen as follows.
The first experiment is done at $N=10,30,100,300,1000$ for constant multimodality parameters $A=3$ and $\alpha=2\pi$.
Hence, $f_\mathrm{ras}$ becomes more difficult to optimize with increasing $N$.
For the second experiment, one varies $N$ and $A$ together as
$(N, A) = (10,65), (30,33), (100,12), (300,7),(1000,3)$ (at $\alpha=2\pi$).
Large $A$ are chosen for small $N$ and vice-versa.
The values are chosen such that CSA~\cref{eq:sqrtN} yields roughly a constant success rate $P_S\approx0.9$ at $\mu=\mumax$.
The test provides ``similar difficulty" for an ES operating at constant population size and will serve as a reference.

For the initialization, one chooses $\vb{y}^{(0)}=2\ceil{\alpha A/2}\vb{1}$ (outside the local attraction region), $\sigma^{(0)} = \signzero\norm{\vb{y}^{(0)}}/N$ via \cref{eq:sign} ensuring a large initial step-size, and $\vb{s}^{(0)}=\vb{1}$ for the CSA.
The runs terminate for $f<f_\mathrm{stop}=10^{-3}$ (global convergence) or $\sigma<10^{-3}$ with $f\geq f_\mathrm{stop}$ (local convergence).
For each parameter set, 50 trials are evaluated and the success rate $P_S$ is measured.
We measure the expected runtime $E_r$ (in function evaluations) to reach global convergence by including the number of function evaluations of successful ($F_s$) and unsuccessful ($F_u$) runs. For $P_S>0$, it is estimated as \cite{AH05a}
\begin{equation}\label{eq:ert}
	E_r = (F_s+F_u)/P_S.
\end{equation}
For the PCS, we choose the following parameters (see Sec.~\cref{sec:sph_ran})
\begin{subequations}
\begin{align}
	&\text{APOP: CSA~\cref{eq:sqrtN} and P2 (Tab.~\cref{tab:configs})}  \label{apop} \\
	&\text{pcCSA: CSA~\cref{eq:sqrtN} and P2 (Tab.~\cref{tab:configs})}  \label{pccsa} \\
	&\text{PSA: CSA~\cref{eq:han} and P2 (Tab.~\cref{tab:configs})}. \label{psa}
\end{align}
\end{subequations}
The population size limits are $\mu^{(0)}=\mu_\mathrm{min}=4$ and $\mumax=1024$.
There are two reasons for limiting $\mu$.
One reason are limited computational resources, especially when optimizing Rastrigin at large $N=1000$.
The second reason is that the PCS are compared to constant $\mu=\mumax$ runs.
Certain examples will show that $\mumax$ is not reached by the PCS, even though it should be reached to achieve better performance.

\begin{figure}[t]
	\centering
	\begin{subfigure}{\columnwidth}
		\centering
		\includegraphics[width=\mywidth]{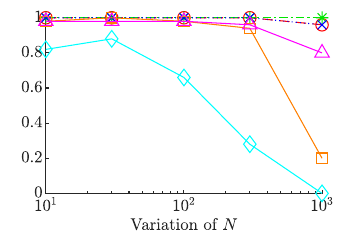}
		\includegraphics[width=\mywidth]{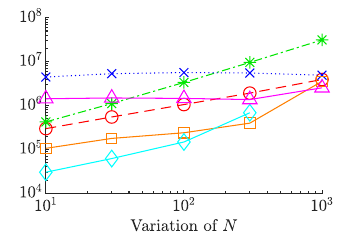}
		\caption{Variation of $N$ at $A=3$ and $\alpha=2\pi$.}
		\label{fig:ras_a}
	\end{subfigure}
	\begin{subfigure}{\columnwidth}
		\centering
		\includegraphics[width=\mywidth]{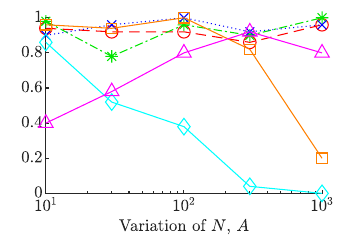}
		\includegraphics[width=\mywidth]{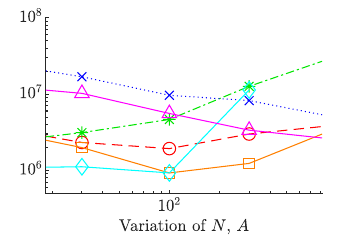}
		\caption{Variation of $N$ and $A$ ($\alpha=2\pi$).}
		\label{fig:ras_b}
	\end{subfigure}
	\caption{Success rate $P_S$ (left column) and expected runtime $E_r$ (right column) on $f_\mathrm{ras}$.
		The CSA with $\mu=\mumax$ are shown as $\circ$ \cref{eq:sqrtN}, $*$ \cref{eq:linN}, and $\times$ \cref{eq:han}.
		The PCS are shown as $\square$ \cref{apop}, $\diamond$ \cref{pccsa}, and $\triangle$ \cref{psa}.}
	\label{fig:ras}
\end{figure}

The results are shown in Fig.~\cref{fig:ras} in terms of $P_S$ and $E_r$.
The following observations can be made.
At constant $\mu=\mumax$, CSA~\cref{eq:sqrtN} yields a very good overall performance.
While $P_S$ is comparable to CSA~\cref{eq:linN} and \cref{eq:han}, its $E_r$ levels are the lowest (only exception is $N=10$ in Fig.~\cref{fig:ras_b}).
While $E_r$ remains approximately constant in Fig.~\cref{fig:ras_b}, CSA~\cref{eq:linN} and \cref{eq:han} show notable upward and downward trends for $E_r$, respectively, due to their different adaptation characteristics.
With active $\mu$-control, APOP and pcCSA stay mostly below $\mu=\mumax$ of CSA~\cref{eq:sqrtN} (the CSA they utilize).
This is important as it illustrates a more effective search with active $\mu$-control.
The highest efficiency (lowest $E_r$) for small $N$ are achieved by pcCSA, which is related to very low $\mu$-levels in the sphere limits (see Tab.~\cref{tab:sqrtN}).
However, the pcCSA shows a significant drop in $P_S$ (and increase of $E_r$) for large $N$.
The APOP shows very good overall results, showing relatively high $P_S$-levels and satisfactory $E_r$ for the tested configurations.
However, its performance also decreases to some extent for large $N$.
The PSA shows lower $E_r$-values than its CSA~\cref{eq:han} at $\mu=\mumax$, which is good.
In Fig.~\cref{fig:ras_b}, its $E_r$-slope closely follows CSA~\cref{eq:han} which illustrates that the CSA properties are recovered with active population control.
In contrast to APOP and pcCSA, the PSA performs best at large $N=1000$.
At small $N=10$, it tends to pick up more fluctuations of $\sigma$ that falsely indicate good performance when measuring $\ppt$ (similar to the bottom plot in Fig.~\cref{fig:psa_dyn}).
At large $N$, the fluctuations of the contributing terms $\ppm$ and $\ppc$ are reduced and the detection of insufficient performance becomes more robust.
Exemplary dynamics of the PCS on $f_\mathrm{ras}$ are given in the supplementary material \cref{ap:add_dyn}.

\begin{figure}[t]
	\centering
	\begin{subfigure}{\columnwidth}
		\centering
		\includegraphics[width=\mywidth]{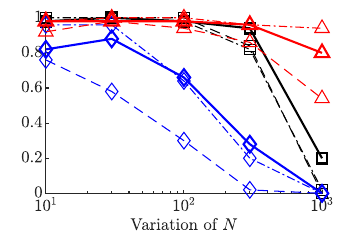}
		\includegraphics[width=\mywidth]{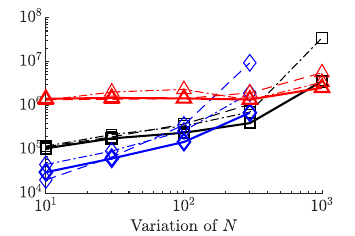}
		\caption{Variation of $N$ at $A=3$ and $\alpha=2\pi$.}
		\label{fig:ras_resig_a}
	\end{subfigure}
	\begin{subfigure}{\columnwidth}
		\centering
		\includegraphics[width=\mywidth]{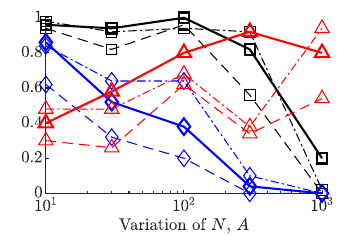}
		\includegraphics[width=\mywidth]{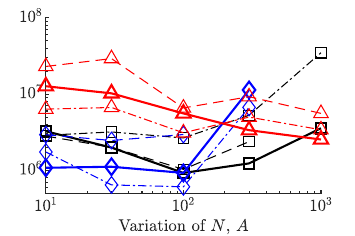}
		\caption{Variation of $N$ and $A$ ($\alpha=2\pi$).}
		\label{fig:ras_resig_b}
	\end{subfigure}
	\caption{Setup as shown in Fig.~\cref{fig:ras}.
	The PCS are shown as $\square$ \cref{apop}, $\diamond$ \cref{pccsa}, and $\triangle$ \cref{psa}.
	Given P2 in Tab.~\cref{tab:configs}, rescaling $r_\sigma$ is varied using \cref{eq:r2} (solid, bold, same data as in Fig.~\cref{fig:ras}), \cref{eq:r3} (dash-dotted), and \cref{eq:r1} (dashed).
	}
	\label{fig:ras_resig}
\end{figure}

Figure~\cref{fig:ras_resig} shows the same experiment as in Fig.~\cref{fig:ras}, now with $\sigma$-rescaling $r_\sigma$ being varied for each of the three PCS.
For APOP one observes improved $P_S$ (left) and $E_r$ (right) for \cref{eq:r2} most of the time.
For pcCSA mixed results are obtained.
Either \cref{eq:r2} or \cref{eq:r3} yields higher $P_S$ and lower $E_r$, respectively.
Similar (mixed) results are observed for PSA.
Linear scaling improves $E_r$ for small $N$ in Fig.~\cref{fig:ras_resig_b}, but not at large $N$.
Recall that linear scaling was originally used by \cite{NA18} (see also discussion of \cref{eq:sigma_rescale_linMU}).
In nearly all cases, no rescaling \cref{eq:r1} yields worse results for all PCS.
%The overall characteristics of the PCS (bold lines) are mostly preserved.
The stability tests with predefined $\mu$-schedule in Sec.~\cref{sec:schedule} have shown clear advantages of \cref{eq:r2}, which is not observed in Fig.~\cref{fig:ras_resig}.
Using active PCS introduces feedback that controls $\mu$ based on the measured $\Per$.
The realized changes in $\mu$ (over multiple generations) are usually moderate compared to a fixed schedule. 
This leads to smaller $\sigma$-changes due to rescaling.
Furthermore, the waiting time $\Delta_g>0$ helps to stabilize the $\sigma$-dynamics for any $r_\sigma$, see also Fig.~\cref{fig:schedule_wait}.
Experiments on $f_\mathrm{sph}$ and $f_\mathrm{ran}$ with varying $r_\sigma$ (not shown) yield similar inconclusive results.
The only exception is the PSA which (on the sphere) appears more susceptible to larger changes of \cref{eq:r3} due to its aggregation \cref{eq:psa_sigmachange}.

	\section{Conclusions and Outlook}
\label{sec:conc}

In this paper, adaptive (online) population control strategies (PCS) were investigated for a multi-recombinative $\muilam{\mu}{\lambda}$-ES with cumulative step-size adaptation (CSA) and isotropic mutations.
To this end, the adaptation properties of standard CSA-variants were discussed on the sphere as a function of population size $\mu$ and dimensionality $N$.
Furthermore, scaling laws for the generation number ($\sqrt{N}$-law, independent of $\mu$)
and rescaling of the mutation strength ($\sqrt{\mu}$-law) were derived on the sphere in the limit of large population sizes.

The obtained scaling laws were implemented into three state-of-the-art PCS, namely APOP \cite{NH17}, pcCSA \cite{HB16}, and PSA \cite{NA18}.
%The experimental (and theoretical) analysis of individual PCS and the comparison among each other has shown significant dependence of their respective performance measures on the CSA-implementation.
The experimental (and theoretical) analysis of the PCS has shown significant dependence of their respective performance measures on the CSA-variants.
On the selected test bed (sphere, random, and Rastrigin functions), APOP and pcCSA using CSA~\cref{eq:sqrtN} (cumulation constant $= 1/\sqrt{N}$, damping $= \sqrt{N}$) perform well due to comparably fast $\sigma$-adaptation.
This improves overall progress and also helps the PCS to discern good from insufficient ES-performance.
PSA works more efficiently using CSA~\cref{eq:han} due to its underlying cumulation paths.
Especially on the Rastrigin function, the PCS benefit from short data aggregation periods ($\sqrt{N}$-law) together with equally short waiting times for the decorrelation of $\mu$-dependent data.
For $\sigma$-rescaling $r_\sigma$, stability tests on the sphere have shown clear advantages of a $\sqrt{\mu}$-law.
However, with active PCS the effects of $r_\sigma$ are less pronounced.
Strengths and weaknesses of the individual PCS have been discussed.
In conclusion, there is no clear winner among the PCS.
Best overall performance was achieved by APOP, providing good results across all parameter variations.
On the other hand, pcCSA was the most efficient PCS at low $N$, while PSA achieved the best results at large $N$.

Of course, the performance of individual PCS can be improved by selective parameter tuning.
However, as the aim of the present work was to investigate basic properties of CSA and PCS, this is left for future research.
Further research aims at testing favorable PCS configurations on a larger benchmark set of noisy and multimodal functions.
The goal is to have good performance over a large dimensionality range, while maintaining efficiency on unimodal functions.
Another potential line of research is the improvement (or hybridization) of different population control methods, which may evaluate the ES performance in both search and fitness space.
Future research should also investigate the interaction of PCS with active covariance matrix adaptation (CMA).
These investigations should, hopefully, yield improved algorithms that perform well on a variety of test functions and also scale well with the dimensionality.

	% --- TEXSTUDIO: txs:///compile | txs:///bibtex | txs:///view ---
	%\bibliography{oa.bib,hgb_my.bib,hgb_mylit.bib}
	% ===> generates *.bbl file
	
	% --- build: TEXSTUDIO: txs:///compile | txs:///view ---
	% Generated by IEEEtran.bst, version: 1.14 (2015/08/26)

	\bibliographystyle{IEEEtran}
	%\bibliography{oa.bib, hgb_my.bib, hgb_mylit.bib}
	
	\numberwithin{equation}{section}
	%\clearpage
	\appendices
		\section{Supplementary Material}

\subsection{Mutation Strength Rescaling}
\label{ap:sigma_rescale}

In this section, the rescaling of the mutation strength $\sigma$ of the PSA from \cite{NA18} is analyzed.
It will be shown that for sufficienly large $N$ it corresponds to result \cref{eq:sigma_rescale_linMU}, i.e., a linear relation between mutation strengths and population sizes.
The $\sigma$-change introduced in \cite[(17)]{NA18}
agrees with the form $\sigma^{(g+1)} = r_\sigma\sigma^{(g)}$ of \cref{sec:gen_num_sig}.
The authors use an optimal $\sigma^*$ derived from a quality gain analysis on the sphere as
\begin{align}\begin{split}
		\label{eq:time_changeSig_2}
		\sigma^* = \frac{c \mu_w N}{N-1+c^2\mu_w}.
\end{split}\end{align}
A few remarks on \cref{eq:time_changeSig_2} are given.
Note that $\mu_w=\mu$ for intermediate recombination.
Furthermore, the coefficient $c$ is given as the sum
$c= -\sum_{i=1}^\lambda w_i \EV{\mathcal{N}_{i;\lambda}} = -\frac{1}{\mu}\sum_{i=1}^\mu \EV{\mathcal{N}_{i;\lambda}}$ for $w_i=1/\mu$ ($1 \leq i \leq \mu$) and $w_i=0$ ($\mu < i \leq \lambda$).
The coefficient $c$ can be related to the progress coefficient $\CMULAM$ using \cite[(D.33)]{Arn02}. 
Assuming zero noise and the case of minimization, one has ($h^{1,0}_{\mu,\lambda}=\CMULAM$), such that
\begin{align}\begin{split}
		\label{eq:time_changeSig_c}
		c = -\frac{1}{\mu}\sum_{i=1}^\mu \EV{\mathcal{N}_{i;\lambda}} = -\frac{1}{\mu}\sum_{i=1}^\mu \EV{z_{i;\lambda}} = \CMULAM.
\end{split}\end{align}
Hence, \cref{eq:time_changeSig_2} yields for intermediate recombination 
\begin{align}\begin{split}
		\label{eq:time_changeSig_3}
		\sigma^* = \frac{\CMULAM \mu N}{N-1+\CMULAM^2\mu} \overset{N\rightarrow\infty}{\simeq} \CMULAM\mu.
\end{split}\end{align}
The result \cref{eq:time_changeSig_3} yields the optimal $\hat{\sigma}^* = \CMULAM\mu$ for $N\rightarrow\infty$ and $\mu\ll N$, which agrees with the $\sigma^*$ maximizing \cref{eq:sph_Ninf}.
These results correspond to a rescaling of $\sigma$ linear with the population size, as shown in \cref{eq:sigma_rescale_linMU}.
However, the limit $\mu\rightarrow\infty$ ($N<\infty$) yields the asymptotic
$\frac{\CMULAM \mu N}{N-1+\CMULAM^2\mu} \simeq \frac{N}{\CMULAM}$, which does not recover the square root law derived in \cref{eq:sigma_rescale_sqrtMU}.

\subsection{PSA-CSA-ES Analysis}
\label{ap:psa_analysis}

In this section, the analysis of the simplified PSA-CMA-ES from \cite{NA18} is shown.
To this end, only the CSA-ES is considered without covariance matrix adaptation.
Furthermore, intermediate recombination is used.
These assumptions, along with the normalization in \cref{eq:psa_ss_exp2}, will allow to derive the corresponding cumulation paths \cref{eq:csa_pm} and \cref{eq:psa_sigmachange} of the PSA.

The main idea of the PSA is to define two cumulation paths for the change of the mean search vector $\vb{y}^{(g)}$ and the covariance matrix $\vb{C}^{(g)}$.
One measures $\dm^{(g+1)} = \vb{y}^{(g+1)} - \vb{y}^{(g)}$ for the positional change.
For the change of the covariance matrix $\vb{C}$ and the mutation strength $\sigma$, one defines $\bm{\Sigma}^{(g)}=(\sigma^{(g)})^2\vb{C}^{(g)}$ first.
Then, one evaluates the difference $\dS^{(g+1)} = (\sigma^{(g+1)})^2\vb{C}^{(g+1)} - (\sigma^{(g)})^2\vb{C}^{(g)}$. 
Furthermore, $\dm^{(g+1)}$ and $\dS^{(g+1)}$ are transformed using their respective Fisher transformation matrices (details in \cite{NAO23}) to achieve invariance w.r.t.~the chosen (normal) search space distribution.
The transformed quantities are \cite[(15a),(15b)]{NAO23}
\begin{align}
	\widetilde{\Delta}_m^{(g+1)} &= \sqrt{\bm{\Sigma}^{(g)}}^{-1}\Delta\vb{m}^{(g+1)} \label{ap:psa_ss_exp3a} \\
	%&\coloneqq \mathcal{I}_m^{1/2}\qty[\Delta\vb{m}] = \sqrt{\bm{\Sigma}}^{-1}\Delta\vb{m} \label{ap:psa_ss_exp3a} \\
	\widetilde{\Delta}_c^{(g+1)} &= \frac{1}{\sqrt{2}} 
	\operatorname{vec}
	\qty(\sqrt{\bm{\Sigma}^{(g)}}^{-1}
	\Delta\bm{\Sigma}^{(g+1)}
	\sqrt{\bm{\Sigma}^{(g)}}^{-1}). \label{ap:psa_ss_exp3b} %\mathcal{I}_c^{1/2}\qty[\Delta\bm{\Sigma}] 
\end{align}
The inverse matrix square root is denoted by $\sqrt{\bm{\Sigma}}^{-1}$ (see also \cite{hansen2023cma} for more details on the transformation).
Furthermore, $\mathrm{vec()}$ denotes the transformation of the argument (matrix) into a vector form for the cumulation path.
Using cumulation constant $\beta$, the PSA measures the two update paths (cf.\ \cite[(12)]{NA18})
\begin{align}
	\gmcm{\vb{p}}{g+1} &= (1-\beta)\gmcm{\vb{p}}{g} + \sqrt{\beta(2-\beta)/E_F}
	\widetilde{\Delta}_m^{(g+1)} \label{ap:psa_ss_exp1a} \\
	\gmcc{\vb{p}}{g+1} &= (1-\beta)\gmcc{\vb{p}}{g} + \sqrt{\beta(2-\beta)/E_F}\widetilde{\Delta}_c^{(g+1)}. \label{ap:psa_ss_exp1b}
\end{align}
The quantity $E_F$ includes the expected value of the Fisher-transformed updates $\widetilde{\Delta}_m$ and $\widetilde{\Delta}_c$, see \cite[(13)]{NA18} and serves as a normalization factor.
For inactive matrix adaptation, only the first two terms are relevant giving 
$E_F\approx \frac{N}{\mu} + \frac{2N(N-E_\chi^2)}{E_\chi^2}\gamma_\sigma^{(g+1)}\frac{c_\sigma^2}{d_\sigma^2}$ (where $\gamma_\sigma^{(g+1)}\rightarrow1$ is a time-dependent quantity quickly approaching one and $E_\chi^2\propto N$).
For sufficiently small ratios $c_\sigma^2/d_\sigma^2\ll1$ (which is usually satisfied for $c_\sigma^{-1}\propto d_\sigma$), the second term is negligible and we can approximate
\begin{align}\label{eq:psa_ss_exp2}
	E_F \approx N/\mu.
\end{align}
The PSA measures the squared norm of both contributions \cref{ap:psa_ss_exp1a} and \cref{ap:psa_ss_exp1b}.
Aggregating both vectors into a single update vector $\vb{p}_\theta^{(g+1)} = (\gmcm{\vb{p}}{g+1}, \gmcc{\vb{p}}{g+1})$, one evaluates
\begin{equation}\label{ap:psa_components}
	\ppt = \ppm + \ppc.
\end{equation}
The result \cref{ap:psa_components} is discussed below \cref{eq:psa_components}.

Now the simplified evolutions paths are evaluated.
Note that without covariance matrix adaptation, one has 
$\bm{\Sigma} = \sigma^2\vb{I}$ with inverse square root $\sqrt{\bm{\Sigma}}^{-1} = \frac{1}{\sigma}\vb{I}$.
The position update yields $\vb{y}^{(g+1)} = \vb{y}^{(g)} + \frac{1}{\mu}\SUMM \vb{x}_{m;\lambda}$ with $\vb{x}_{m;\lambda}$ denoting the selection of the $m=1,...,\mu$ best mutations, see Alg.~\cref{alg:pcs_algos}.
Hence, one obtains for \cref{ap:psa_ss_exp3a} with $\vb{x} = \sigma\vb{z}$ and $\vb{z}\sim\mathcal{N}(\vb{0},\vb{1})$
\begin{align}\begin{split}\label{ap:psa_ss_exp6}
	\widetilde{\Delta}_m^{(g+1)} &= \frac{1}{\sigma^{(g)}}\vb{I}\qty(\vb{y}^{(g+1)} - \vb{y}^{(g)}) \\
	&= \frac{1}{\mu}\SUMM \vb{z}_{m;\lambda} = \rec{\vb{z}}^{(g+1)}.
\end{split}\end{align}
By inserting \cref{ap:psa_ss_exp6} and \cref{eq:psa_ss_exp2} into \cref{ap:psa_ss_exp1a}, one gets the same form as the CSA cumulation path \cref{eq:csa_s} with additional normalization factor $1/\sqrt{N}$
\begin{align}\label{ap:csa_pm}
	\gmcm{\vb{p}}{g+1} &= (1-\beta)\gmcm{\vb{p}}{g} + \sqrt{\beta(2-\beta)\mu/N} \rec{\vb{z}}^{(g+1)}. 
\end{align}
Now \cref{ap:psa_ss_exp3b} is evaluated using $\sqrt{\bm{\Sigma}}^{-1} = \frac{1}{\sigma}\vb{I}$.
One gets
\begin{align}\begin{split}\label{ap:psa_ss_exp7}
		&\widetilde{\Delta}_c =  \frac{1}{\sqrt{2}}
		\operatorname{vec}\qty(
		\frac{1}{\sigma^{(g)}}\vb{I}\qty[(\sigma^{(g+1)})^2\vb{I} - (\sigma^{(g)})^2\vb{I}]\frac{1}{\sigma^{(g)}}\vb{I} ) \\
		&= \frac{1}{\sqrt{2}}
		\operatorname{vec}\qty(
		\qty[
		\frac{(\sigma^{(g+1)})^2}{(\sigma^{(g)})^2} - 1]\vb{I})
		%&= \frac{1}{\sqrt{2}}\qty[\frac{(\sigma^{(g+1)})^2}{(\sigma^{(g)})^2} - 1]  \operatorname{vec}(\vb{I}) 
		= 
		\frac{1}{\sqrt{2}}\qty[\frac{(\sigma^{(g+1)})^2}{(\sigma^{(g)})^2} - 1] \vb{1}.
\end{split}\end{align}
The notation $\vb{1}=[1,\dots,1]$ (with length $N$) was used.
For the vectorization of the identity, one may use $\operatorname{vec}(\vb{I})=\vb{1}$ by considering only the non-zero elements since the aggregation of constant zero values is useless.
Hence, the update rule yields
%Our simplified PSA has only $N$ degrees of freedom (mutation strength acting on $N$ components), which is in contrast to the full covariance matrix adaptation having \cite{NA18}
%
\begin{align}\begin{split}\label{ap:psa_sigmachange}
		\gmcc{\vb{p}}{g+1} = (1-\beta)\gmcc{\vb{p}}{g} + \sqrt{\frac{\beta(2-\beta)\mu}{(2N)}}\qty[\frac{(\sigma^{(g+1)})^2}{(\sigma^{(g)})^2}-1]\vb{1}.
\end{split}\end{align}

\subsection{Sphere Steady-State of the PSA-CSA-ES}
\label{ap:psa_stst}

\begin{figure}[t]
	\centering
	\begin{subfigure}{\columnwidth}
		\centering
		\includegraphics[width=\mywidth]{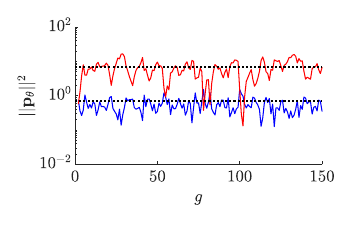}
		\includegraphics[width=\mywidth]{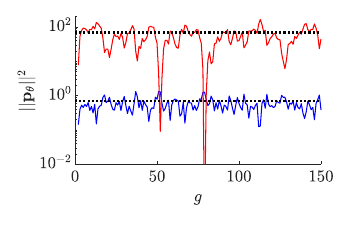}
		\caption{$N=10$ with $\mu=100$ (left) and $\mu=1000$ (right).}
		\label{fig:psa_stst_N10}
	\end{subfigure}
	\begin{subfigure}{\columnwidth}
		\centering
		\includegraphics[width=\mywidth]{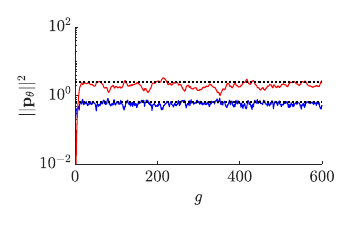}
		\includegraphics[width=\mywidth]{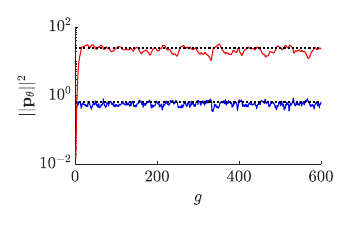}
		\caption{$N=100$ with $\mu=100$ (left) and $\mu=1000$ (right).}
		\label{fig:psa_stst_N100}
	\end{subfigure}
	\caption{PSA-CSA-ES at constant population size on the sphere, using CSA~\cref{eq:new_han_v1} and $\beta=1/\sqrt{N}$.
	The blue signal shows $\ppm$ and the red signal $\ppc$, respectively (the sum $\ppt$ is not shown to improve visibility).
	The respective black horizontal lines show steady-state predictions $\ppm$ \cref{eq:csa_ss_13d} and $\ppc$ \cref{eq:psa_ss_pc2}, both evaluated at $\gamma=0.88$ (see Fig.~\cref{fig:compare_CSAs_gamma}).
	}
	\label{fig:psa_stst}
\end{figure}

In this section, sphere steady-state contributions of $\ppm$ and $\ppc$ (see \cref{eq:psa_components} or \cref{ap:psa_components}) are derived by assuming a constant population size $\mu$.
As already mentioned, the cumulation path of $\ppm$ works analogously to the cumulation of the CSA-ES (cf.~\cref{eq:csa_pm} and \cref{eq:csa_s})
Hence, one may directly apply the steady-state derivation of the CSA from \cite{OB24b} to the derivation of $\ppm$.
Only the main derivation steps are recalled now.
The derivation requires the evaluation of $||\gmcm{\vb{p}}{g+1}||^2$.
Assuming that progress on the sphere is achieved along a certain component $A$ (unit vector $\gvec{e}{g}_A$) with the remaining $N-1$ components being selectively neutral, a second equation of $\gmcm{\vb{p}}{g+1}$ along direction $A$ is derived, such that one has to evaluate the scalar product
\begin{align}\begin{split}\label{eq:csa_ss_4}
		p_A^{(g+1)} &= \gmcm{\vb{p}}{g+1}\gvec{e}{g+1}_A.
\end{split}\end{align}	
In order to obtain closed-form solutions of the steady-state, approximations are applied in \cite{OB24b} by assuming sufficiently small progress rates.
Then, one imposes the steady-state conditions (in expectation) by setting $\EV{\norm{\gmcm{\vb{p}}{g+1}}^2}=\norm{\gmcm{\vb{p}}{g}}^2=\norm{\vb{p}_m}^2$ and 
$\EV{p_A^{(g+1)}} = p_A^{(g)} = p_A$.
One obtains expected values over the selected mutation direction, i.e., $\EV{||\rec{\vb{z}}||^2}$ and $\EV{\rec{z_A}}$ with $\rec{z_A} = \rec{\vb{z}}\vb{e}_A$.
Accounting for the fact that the cumulation constant of PSA is $\beta$ (instead of $\cs$) and the normalization of the cumulation path includes $\sqrt{\mu/N}$ (instead of only $\sqrt{\mu}$), one obtains the steady-state equation for $p_A$ (analogous equation to \cite[(30)]{OB24b}) as 
\begin{align}\begin{split}\label{eq:csa_ss_11}	
		p_A &= \sqrt{\frac{\beta(2-\beta)\mu}{N}}\frac{\EV{\rec{z_A}}-\frac{\sigma^{*}}{N}\EV{||\rec{\vb{z}}||^2}}
		{\beta+(1-\beta)\frac{\sigma^{*}}{N} \EV{\rec{z_A}}}.
\end{split}\end{align}	 
Similarly, one gets for the squared norm the steady-state equation (cf.~\cite[(31)]{OB24b})
\begin{align}\begin{split}\label{eq:csa_ss_12}
		||\vb{p}_m||^2 &=\frac{\mu}{N}\EV{||\rec{\vb{z}}||^2} \!+\!\frac{2(1-\beta)}{\beta(2-\beta)}
		\sqrt{\frac{\beta(2-\beta)\mu}{N}}p_A\EV{\rec{z_A}} 
\end{split}\end{align}
Now \cref{eq:csa_ss_11} can be inserted into \cref{eq:csa_ss_12}, which yields
\begin{align}\begin{split}\label{eq:csa_ss_13}
		||\vb{p}_m||^2 &= \frac{\mu}{N}\EV{||\rec{\vb{z}}||^2} \\
		&\quad+ \frac{2(1-\beta)\mu}{N}
		\frac{\EV{\rec{z_A}}^2 - \frac{\sigma^{*}}{N}\EV{\rec{z_A}}\EV{||\rec{\vb{z}}||^2}}
		{\beta+(1-\beta)\frac{\sigma^{*}}{N}\EV{\rec{z_A}}}.
\end{split}\end{align}
The expected values in \cref{eq:csa_ss_13} are known quantities.
One has
\begin{subequations}
	\begin{align}
		%\varphi^* &= \sqrt{2N}\CTHETA - \sigma^{*2}/2\mu \qq{from \cref{sec:dyn_phi_large}} \label{eq:csa_ss_large1} \\
		\begin{split}
		\EV{\rec{z_A}} &= \sqrt{2N}\CTHETA/\sigma^*, \qq{from \cite[(5.7)]{Arn02},} \\
		&\hspace{-5em}\qq{with $\sqrt{1+\sigma^{*2}/2N} \simeq \sigma^*/\sqrt{2N}$, $\CMULAM\simeq\CTHETA$} \label{eq:csa_ss_large2}\end{split} \\
		\EV{||\rec{\vb{z}}||^2} &= N/\mu, \qq{from \cite[(5.2)]{Arn02}.} \label{eq:csa_ss_large3}
	\end{align}
\end{subequations}
Hence, \cref{eq:csa_ss_13} can be evaluated as
\begin{align}\begin{split}\label{eq:csa_ss_13b}
		||\vb{p}_m||^2 = 1 + 2(1-\beta)
		\frac{\frac{2\CTHETA^2\mu}{\sigma^{*2}} - \sqrt{2/N}\CTHETA}
		{\beta+(1-\beta)\sqrt{2/N}\CTHETA}.
\end{split}\end{align}
Result \cref{eq:csa_ss_13b} is a function of the normalized mutation strength $\sigma^*$.
Therefore, the steady-state $||\vb{p}_m||^2$ depends on the CSA-parametrization.
It is now useful to apply assumption \cref{eq:signzero_approx_gam_v2} to express the steady-state $\sigma^*=\signss(\gamma)$ as a function of scaling parameter $\gamma$, such that one gets
\begin{align}\begin{split}\label{eq:csa_ss_13c}
		||\vb{p}_m||^2 &= 1 + 2(1-\beta)
		\frac{\frac{2\CTHETA^2\mu}
		{\gamma^2(8N)^{1/2}\CTHETA\mu} 
		- \sqrt{2/N}\CTHETA}
		{\beta+(1-\beta)\sqrt{2/N}\CTHETA} \\
		&= 1 + 2(1-\beta)
		\frac{\sqrt{2/N}\CTHETA\qty(\frac{1}{2\gamma^2}-1)}
		{\beta+(1-\beta)\sqrt{2/N}\CTHETA}, 
\end{split}\end{align}
such that after rearranging one finally gets
\begin{align}\begin{split}\label{eq:csa_ss_13d}
	||\vb{p}_m||^2 &=  1 - \frac{2-1/\gamma^2}{1+\frac{\beta}{\sqrt{2/N}\CTHETA(1-\beta)}}.
\end{split}\end{align}
For sufficiently slow adaptation, i.e., in the limit $\gamma\rightarrow1$, \cref{eq:csa_ss_13d} is expected to yield $||\vb{p}_m||^2<1$ (see also Fig.~\cref{fig:psa_stst}). 

Now the steady-state of $\ppc$ is derived.
For brevity the following notation is introduced for the relative $\sigma$-change derived in \cref{ap:psa_sigmachange}
\begin{align}\label{eq:psa_ss_sDef}
	s \coloneqq \qty(\sigma^{(g+1)}/\sigma^{(g)})^2 - 1.
\end{align}
The cumulation path yields ($\vb{1}=[1,\dots,1]$, $\sum_i \vb{1} = N$)
\begin{align}\label{eq:psa_ss_exp20a}
	\gmcc{\vb{p}}{g+1} &= (1-\beta)\gmcc{\vb{p}}{g} + \sqrt{\frac{\beta(2-\beta)\mu}{2 N}}
	s \vb{1}.
\end{align}
\mycomment{and its squared norm
\begin{align}\begin{split}
	\label{eq:psa_ss_exp20b}
	&\norm{\gmcc{\vb{p}}{g+1}}^2 
	= (1-\beta)^2\norm{\gmcc{\vb{p}}{g} }^2 \\
	&+ 2(1-\beta)\sqrt{\frac{\beta(2-\beta)\mu}{2 N}} s \sum_{i=1}^N (\gmcc{\vb{p}}{g})_i + \frac{\beta(2-\beta)\mu}{2} s^2.
\end{split}\end{align}}
Assuming the steady-state $\vb{p}_c=\gmcc{\vb{p}}{g}=\gmcc{\vb{p}}{g+1}$, \cref{eq:psa_ss_exp20a} yields
\begin{align}\label{eq:psa_ss_exp21a}
	\vb{p}_c &= \frac{1}{\beta}\sqrt{\frac{\beta(2-\beta)\mu}{2 N}}s\vb{1}. %\\ 
	%\sum_{i=1}^N (\vb{p}_c)_i &= \frac{N}{\beta}\sqrt{\frac{\beta(2-\beta)\mu}{2 N}}s.
	%\label{eq:psa_ss_exp21b}
\end{align}
%
%Setting $\norm{\vb{p}_c}^2=\norm{\gmcc{\vb{p}}{g}}^2=\norm{\gmcc{\vb{p}}{g+1}}^2$ and using \cref{eq:psa_ss_exp21b} for the sum in \cref{eq:psa_ss_exp20b}, one has
%
By squaring the result \cref{eq:psa_ss_exp21a}, one gets
\begin{align}\begin{split}\label{eq:psa_ss_exp22}
		%\beta(2-\beta)\norm{\vb{p}_c}^2 &= 2(1-\beta)\sqrt{\frac{\beta(2-\beta)\mu}{2 N}}s \frac{N}{\beta} \sqrt{\frac{\beta(2-\beta)\mu}{2 N}}s \\
		%&\qquad+ \frac{\beta(2-\beta)\mu}{2}s^2 \\
		\norm{\vb{p}_c}^2 &= \qty(\frac{1}{\beta}-\frac12)\mu s^2.
\end{split}\end{align}	
Assuming the steady-state in expectation, the expected value of the result in \cref{eq:psa_ss_exp22} yields
\begin{align}\begin{split}\label{eq:psa_ss_pc}
		\EV{\norm{\vb{p}_c}^2} = \norm{\vb{p}_c}^2  &= \qty(\frac{1}{\beta}-\frac12)\mu\EV{s^2}.
\end{split}\end{align}
As the next step, the relative $\sigma$-change $s$ will be expressed as a function of the progress rate on the sphere.
From \cref{eq:sign} it holds $\sigma^{(g+1)}/\sigma^{(g)} = R^{(g+1)}/R^{(g)}$ for a constant $\sigma^{*,(g+1)}=\sigma^{*,(g)}$ in the sphere steady-state.
One can rewrite $\EV{s^2}$ as
\begin{align}\begin{split}\label{eq:psa_sphereQ_exp1}
		\EV{s^2} &= \EV{\Big(\Big(\ratiog{\sigma}\Big)^2-1\Big)^2} = \EV{\Big(\Big(\ratiog{R}\Big)^2-1\Big)^2} \\
		&= \EV{\Big(\ratiog{R}\Big)^4 - 2\Big(\ratiog{R}\Big)^2+1}.
\end{split}\end{align}
One can now look at the sphere quality gain to obtain an expression for the relation $\EV{(R^{(g+1)}/R^{(g)})^2}$.
The quality gain is defined as the expected value
$q_\mathrm{sph} \coloneqq \EV{(R^{(g)})^2-(R^{(g+1)})^2}$.
By introducing the quality gain normalization $q_\mathrm{sph}^* = q_\mathrm{sph}N/(2R^2)$, see \cite[p.~16]{Arn02}, one has
\begin{align}\begin{split}\label{eq:psa_sphereQ_exp3}
		\EV{(R^{(g+1)}/R^{(g)})^2} = 1-2q_\mathrm{sph}^*/N.
\end{split}\end{align}
The expectation of the fourth order term in \cref{eq:psa_sphereQ_exp1} cannot be easily evaluated, but it can be approximated.
By neglecting fluctuations of the quality gain, one demands $\VAR{x^2} = \EV{x^4}-\EV{x^2}^2 \overset{!}{=} 0$, $x=R^{(g+1)}/R^{(g)}$, such that
$\EV{x^4}=\EV{x^2}^2$.
Inserting \cref{eq:psa_sphereQ_exp3} into \cref{eq:psa_sphereQ_exp1} and using $\EV{(R^{(g+1)}/R^{(g)})^4}=\qty(1-2q_\mathrm{sph}^*/N)^2$, one gets
\begin{align}\begin{split}\label{eq:psa_sphereQ_exp2}
		\EV{s^2} &\approx \EV{\qty(1-\frac{2q_\mathrm{sph}^*}{N})^2 - 2\qty( 1-\frac{2q_\mathrm{sph}^*}{N})+1} \\
		&= \qty(\frac{2}{N}q_\mathrm{sph}^{*})^2 \simeq \qty(\frac{2}{N}\varphi^*)^2.
\end{split}\end{align}
For the last (asymptotic) equality in \cref{eq:psa_sphereQ_exp2}, its was used that $\varphi^*\simeq q_\mathrm{sph}^{*}$ holds for sufficiently large $N$, see \cite[p.~16]{Arn02}.
Since $\varphi^*$ as a function of $\gamma$ is known from \cref{eq:psa_ss_phinorm1}, we are now in a position to insert the results back into  \cref{eq:psa_ss_pc}.
One gets
\begin{align}\begin{split}\label{eq:psa_ss_pc2}
	\norm{\vb{p}_c}^2  &= \qty(\frac{1}{\beta}-\frac12)\mu\qty(\frac{2}{N}\varphi^*)^2 \\
	&= \qty(\frac{1}{\beta}-\frac12)\frac{8\CTHETA^2\mu}{N}(1-\gamma^2)^2.
\end{split}\end{align}
Result~\cref{eq:psa_ss_pc2} is notable since it predicts a linear scaling $\norm{\vb{p}_c}^2\propto\mu$ (assuming $\gamma$ independent of $\mu$).
Furthermore, the contribution of $\norm{\vb{p}_c}^2$ vanishes for the slow adaptation limit $\gamma\rightarrow1$, in which $\sigma$ does not change in expectation.

Figure~\cref{fig:psa_stst} shows steady-state experiments on the sphere to investigate the predictions of $\ppm$ \cref{eq:csa_ss_13d} and $\ppc$ \cref{eq:psa_ss_pc2}.
The horizontal lines show good agreement with the measured signals.
Visually, they do show minor deviations from the mean signal values.
However, since approximations were applied, this was expected.
Furthermore, the goal here is to illustrate basic scaling properties of the PSA-measure w.r.t.~the population size.
As expected, $\ppm<1$ in the sphere steady-state, which is in contrast to $\ppm=1$ during random selection (see also discussion below \cref{eq:psa_components}). 
Furthermore, one observes good agreement for the scaling $\norm{\vb{p}_c}^2\propto\mu$ as $\mu$ is increased (from left to right).
This behavior is a crucial property of the PSA.
For small $\ppt<\Thr$ on the sphere, the PSA increases $\mu$ to increase the respective $\ppc$-contribution.
This usually happens for small initial $\mu$, even though a population increase is not necessary on the sphere.
Hence, the PSA controls $\mu$ to achieve the target value $\ppt\approx\Thr$.
This explains the relatively high $\mu_\mathrm{med}$-levels of the PSA on the sphere observed throughout Sec.~\cref{sec:pcs}.

\subsection{Additional Experiments: APOP}
\label{ap:apop_add}

\begin{figure}[t]
	\centering
	\includegraphics[width=8.8cm]{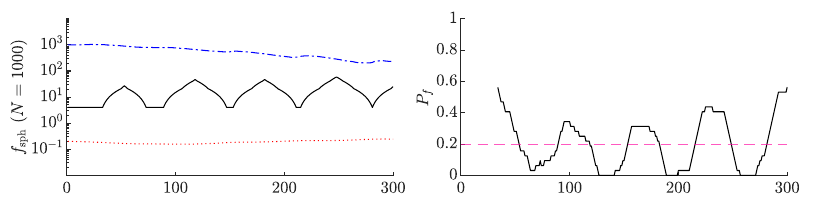}
	\includegraphics[width=8.8cm]{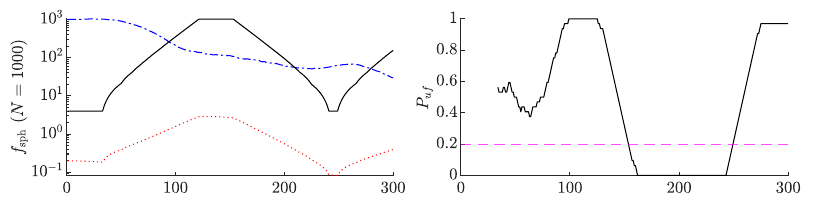}  
	\caption{APOP on the sphere ($N=1000$) with $r_\sigma=1$ (top) and $r_\sigma=\rsigsqrt$ (bottom), using (P1 from Tab.~\cref{tab:configs}).
	}
	\label{fig:apop_rescale}
\end{figure}

In this section, the effects of $\sigma$-rescaling on the performance measure $P_f$ of the APOP are investigated.
The experiments of Fig.~\cref{fig:apop_dyn} (bottom plot) indicate an issue of the APOP with active population control for $r_\sigma=\rsigsqrt$ and no waiting time (P1 from Table~\cref{tab:configs}).
In this case, one observes large oscillations of $\mu^{(g)}$ on the sphere, occurring at larger $N$-values.
The issue is further illustrated in Fig.~\cref{fig:apop_rescale}.
In the given example, the measure $P_{f}$ shows a comparably large initial value of roughly $P_{f}\approx0.5$.
The value is comparably large since the initial population size is small at $\mu^{(0)}=4$.
This is not critical as small populations exhibit more fitness deterioration (see also discussion of Fig.~\cref{fig:new_pcs_deactPop}, top plot).
%The performance measure starts at $P_{f}\approx0.5$ due to a small initial population. 
Since $\Thr=0.2$, $P_{f}>\Thr$ triggers an increase of $\mu$.
While $\mu$ increases and $r_\sigma=1$ (no $\sigma$-rescaling), the top plot indicates performance improvement and $P_{f}$ drops, which is the expected behavior on the sphere.
Small oscillations around $\Thr=0.2$ are observed due to the controlling effect.
In the bottom plot with $\sigma$-rescaling $r_\sigma = \sqrt{\mu^{(g+1)}/\mu^{(g)}}$,
$P_{f}$ increases further indicating worse performance, even though $f$-convergence (blue dash-dotted curve) can be observed.
$P_{f}=1$ at some point, indicating that only fitness deterioration (of the median difference) was measured.
When $\mu$ remains large (and constant) over many generations, $P_{f}$ decreases again and reaches its minimum value zero.
The result are undesired large oscillations of $\mu$ on the sphere.

\begin{figure}[t]
	\centering
	\includegraphics[width=8.8cm]{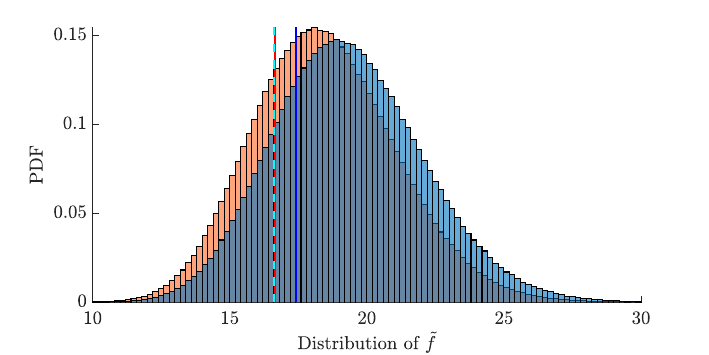}
	\caption{Distribution change of offspring $\tilde{f}$-values on the sphere $N=100$ over two generations $g$ and $g+1$.
		$\muilam{100}{200}$-ES is used with $10^{4}$ repetitions.
		The reddish histogram shows $R^{(g)}=1$ with $\sigma^{(g)}\approx 0.42$. The median over selected values (red vertical line, $\vt=1/2$) yields 16.66.
		The blue histogram shows $R^{(g+1)}=0.98$ and $\sigma^{(g+1)}=\sigma^{(g)}\sqrt{\alpha_\mu}\approx 0.43$ (rescaled using \cref{eq:sigma_rescale_sqrtMU} with $\alpha_\mu=1.05$) with median=17.42 (blue).
		For $R^{(g+1)}=0.98$ with $\sigma^{(g+1)}$ not rescaled, the median yields 16.63 (cyan dashed, histogram not shown).
	}
	\label{fig:apop_hist}
\end{figure}

The observed oscillations are discussed in Fig.~\cref{fig:apop_hist}.
To this end, the distributions of offspring $\tilde{f}$-values are shown together with their corresponding median over the selected values for two consecutive generations.
The goal is to illustrate the effects of $\sigma$-rescaling.
The first experiment at $g$ serves as a reference.
It is initialized randomly at $R^{(g)}=1$ and $\sigma^{(g)}=\gamma\signzero R^{(g)}/N\approx0.42$ was chosen ($\gamma=0.9$) to generate similar conditions as obtained during the dynamic simulation.
Given the parameter configuration at $g$, the ES achieves positive progress using \cref{sec:dyn_phidef} with $\EV{R^{(g+1)}}=0.98$.
Hence, we choose $R^{(g+1)}=0.98$ for the updated histogram.
Since $P_f>\Thr$ is initially large in Fig.~\cref{fig:apop_rescale}, a population size increase is triggered with $\alpha_\mu=1.05$.
Using \cref{eq:sigma_rescale_sqrtMU}, the $\sigma$-rescaling $\sigma^{(g+1)}=\sigma^{(g)}\sqrt{\alpha_\mu}$ is applied to simulate its effect on the APOP.
The distribution of the $\tilde{f}$-values and the median change considerably.
Despite having positive progress, the APOP indicates bad performance via $\fmed{g+1}-\fmed{g}>0$ (minimization).
Hence, the APOP collects information about the change in the distribution of $\tilde{f}$ instead of the actual performance (note that progress rate and quality gain are asymptotically equal on the sphere \cite[p.~16]{Arn02}).
One observes from the histograms that the quality gain expected value and variance change with varying $\sigma$.
Note that if $\sigma$ is not rescaled, i.e., $r_\sigma=1$,
the median at $g+1$ decreases (slightly) compared to $g$, indicating good performance.
Given this observation, the APOP should be evaluated with $r_\sigma=1$ or a waiting time $\Delta_g > 0$ should be introduced.
This is done by choosing P2 (from Tab.~\cref{tab:configs}), which improves the results and reduces the observed oscillations significantly.


\subsection{Additional Experiments: pcCSA}
\label{ap:pccsa_add}

In this section, the failing of the convergence test of the pcCSA (see Fig.~\cref{fig:pccsa_dyn}, bottom) is discussed, giving $P_H>0.05$ consistently on the sphere function.
This leads to an unnecessary increase of $\mu$.

\begin{figure}[t]
	\centering
	\begin{subfigure}{1\columnwidth}
		\centering
		\includegraphics[width=8.8cm]{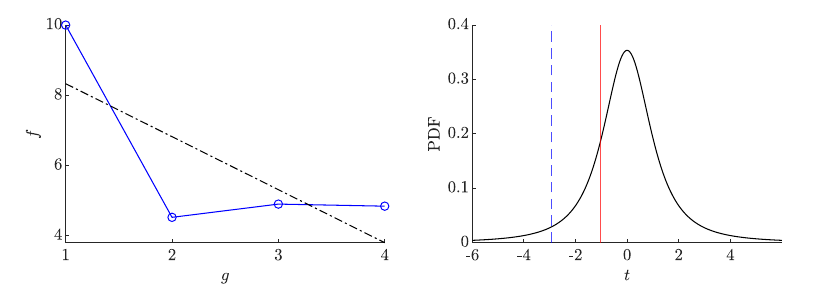} 
		\caption{Hypothesis test of pcCSA at $g=4$ for the sphere at $N=10$ and using $L=\ceil{\sqrt{N}}=4$ (P1 from Tab.~\cref{tab:configs}). 
		On the left, the $f$-values are shown as blue dots and the linear regression line in dash-dotted black.
		On the right, the corresponding $t$-distribution (solid black) is shown with $P_H>\alpha_H$ ($P_H$ in red, threshold 0.05 in  dashed blue).}
		\label{fig:sph_ran_hemi_hyp_a}
	\end{subfigure}
	\begin{subfigure}{1\columnwidth}
		\centering
		\includegraphics[width=8.8cm]{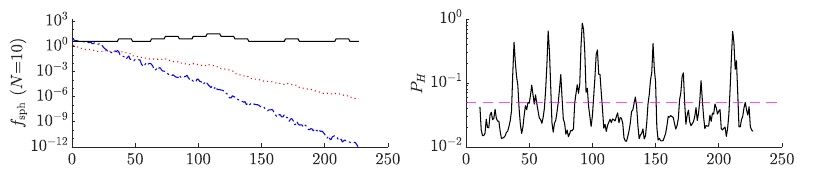}  
		\caption{Dynamics with P2 from Tab.~\cref{tab:configs} using a sample size $L=10$ for the hypothesis test.}
		\label{fig:sph_ran_hemi_hyp_b}
	\end{subfigure}
	\caption{Analysis of pcCSA hypothesis test for small sample size $L$.}
	\label{fig:sph_ran_hemi_hyp}
\end{figure}

To this end, Fig.~\cref{fig:sph_ran_hemi_hyp_a} illustrates the issue of the hypothesis test for a very small sample size $L=4$.
Note that this also occurs at (slightly) larger values of $L$.
One observes a negative slope of $f$-values and the linear regression curve (left).
The hypothesis test on the right side, however, indicates $P_H>0.05$. 
Hence, it fails to reject the null hypothesis of stagnation and the population is increased.
At small $L$ the standard error $s_{\hat{a}}$ attains comparably large values, see \cref{eq:pccmsa_Pval}.
In the case of large errors, the hypothesis test fails to reject $H_0$ and the decision is made towards larger populations, which is the more robust choice.
The results improve significantly for $L=10$ (P2) on the sphere $N=10$ compared to P1, see ``S10" in Tab.~\cref{tab:sqrtN} for pcCSA.
The dynamics of this example are shown in Fig.~\cref{fig:sph_ran_hemi_hyp_b}.

\subsection{Additional Experiments: PSA}
\label{ap:psa_add}

\begin{figure}[t]
	\centering
	\includegraphics[width=8.8cm]{figs/PSA_N10_CSA1_conf2.pdf} 
	\includegraphics[width=8.8cm]{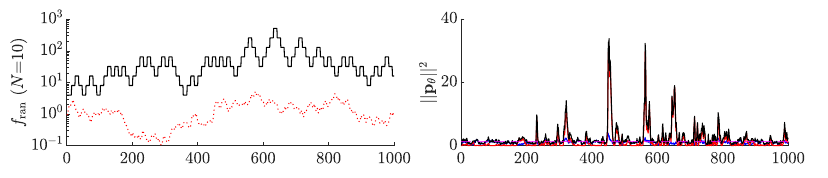} 
	\includegraphics[width=8.8cm]{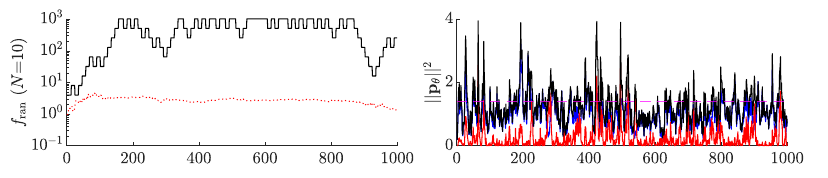} 
	\caption{PSA-dynamics with CSA~\cref{eq:sqrtN} (top, center) and CSA~\cref{eq:han} (bottom) for the random function $N=10$ (P2 from Tab.~\cref{tab:configs}).
	Center and bottom plot use $r_\sigma=1$ (no $\sigma$-rescaling).
	}
	\label{fig:app_psa_dyn}
\end{figure}

Additional experiments of the PSA-CSA-ES are shown on the random function, investigating the influence of $\sigma$-rescaling and suboptimal performance observed in bottom plot of Fig.~\cref{fig:psa_dyn}.
The top plot in Fig.~\cref{fig:app_psa_dyn} shows the bottom plot of Fig.~\cref{fig:psa_dyn} as a reference.
Removing the $\sigma$-rescaling ($r_\sigma=1$ in center plot) does not improve the $\mu$-levels on the random function.
Instead, significantly better performance is obtained using CSA~\cref{eq:han} (bottom plot).
The slower adaptation of $\sigma$ (less fluctuations can be observed) yields a more stable signal of $\ppt$ (less fluctuations of $\ppc$), which in turns leads to better detection of random selection and higher $\mu$-levels.

\subsection{Additional Dynamics}
\label{ap:add_dyn}
\begin{figure}[t]
	\centering
	\includegraphics[width=8.8cm]{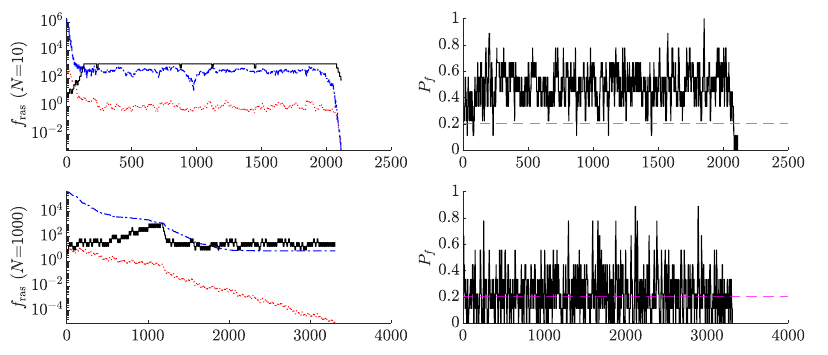}
	\includegraphics[width=8.8cm]{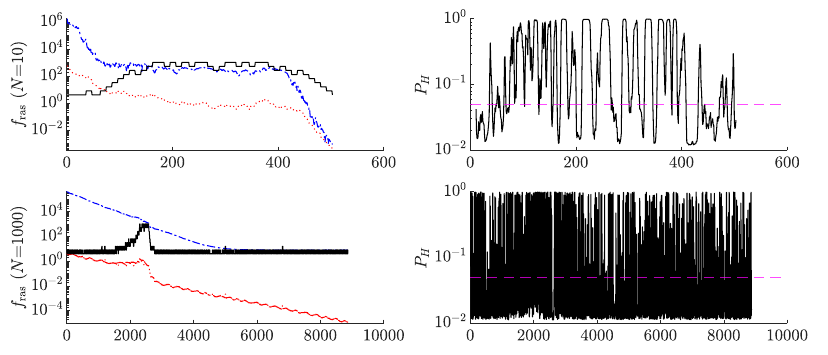}
	\includegraphics[width=8.8cm]{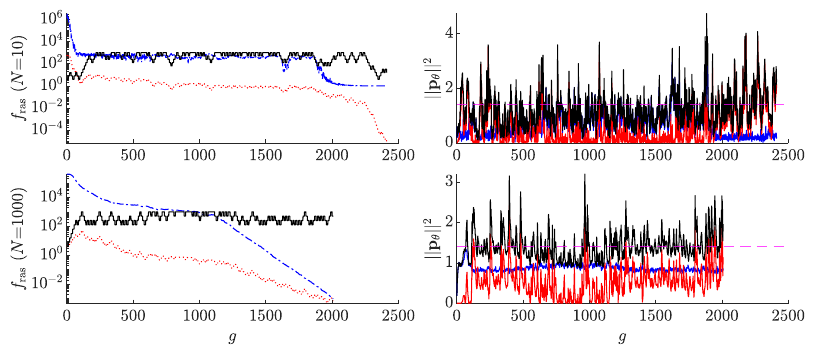}
	\caption{Example runs from Fig.~\cref{fig:ras} for APOP, pcCSA, and PSA, from top to bottom, with each showing $N\!=\!10$, $A\!=\!65$ (top) and $N\!=\!1000$, $A\!=\!3$ (bottom), respectively (see y-axis labels).}
	\label{fig:ras_ex}
\end{figure}
Exemplary dynamics of the experiments from Fig.~\cref{fig:ras} are shown in Fig.~\cref{fig:ras_ex}.
For APOP and pcCSA at $N=10$, one observes good performance.
The $\mu$-levels (solid black) increase and remain high as local attraction is present ($f$-stagnation, dash-dotted blue).
Within the global attractor, $\mu$ decreases again.
For large $N=1000$ the performance of APOP and (especially) pcCSA deteriorate.
During the first phase of $f$-stagnation, one observes a notable increase of $\mu$, which is desired.
However, $\mu$ does not remain stable at high levels. 
On the right one observes that the performance measures $P_f$ and $P_H$ tend to oscillate around their thresholds. 
The PCS do not consistently detect insufficient performance as the ES progresses very slowly, but steadily due to exponentially many local attractors (in $N$).
Hence, both APOP and pcCSA decrease $\mu$ too early (see left plots).
The PSA on the other hand is more robust at large $N=1000$, showing higher $P_S$.
Its $\mu$-levels tend to be comparably high, but its $\mu^{(g)}$-dynamics shows notable oscillations ($N=1000$).
However, note that its $\mu$-levels in the sphere limits of Rastrigin (linear convergence of $f$) are also relatively high.
At small $N=10$, its performance drops to some extent due to larger influence of $\sigma$-fluctuations.

\end{document}